\documentclass[letterpaper]{article} 
\usepackage[preprint]{aaai2027}      
\usepackage[hyphens]{url}            
\usepackage{graphicx}                
\graphicspath{{figures/}}
\usepackage{natbib}                  
\usepackage{caption}                 
\usepackage{algorithm}
\usepackage{algorithmic}

\usepackage{amsmath}
\usepackage{amssymb}
\usepackage{amsfonts}
\usepackage{bm}
\usepackage{booktabs}
\usepackage{multirow}

\newcommand{\R}{\mathbb{R}}

\newcommand{\N}{\mathcal{N}}
\newcommand{\T}{^{\top}}

\newcommand{\spdom}{\Omega}              
\newcommand{\rrr}{\mathbf{r}}            

\newcommand{\zlat}{z}                    
\newcommand{\Wcore}{\mathcal{W}}         
\newcommand{\fieldy}{\mathcal{Y}}        
\newcommand{\obsreg}{\spdom_{\text{obs}}^{(n)}}  
\newcommand{\rstar}{\rrr^{\star}}        
\newcommand{\tstar}{t^{\star}}           

\newcommand{\Obs}{\mathcal{O}}
\newcommand{\Obshist}{\mathcal{O}_{1:n}}
\newcommand{\Datacorpus}{\mathcal{D}}     

\newcommand{\Gobs}{\mathbf{G}}            

\newcommand{\TODO}[1]{}

\title{TRACE: Retrospective Streaming Generation of Physical Fields under Sparse Structured Sensing}

\author{
    Xinyu Zhang\textsuperscript{\rm 1},
    Lihao Chen\textsuperscript{\rm 1},
    Panqi Chen\textsuperscript{\rm 1},
    Lei Cheng\textsuperscript{\rm 1},
    Ting Zhang\textsuperscript{\rm 1},
    Jianlong Li\textsuperscript{\rm 1},
    Shikai Fang\textsuperscript{\rm 1}\thanks{Corresponding author: Shikai Fang \texttt{<fsk@zju.edu.cn>}}
}
\affiliations{
    \textsuperscript{\rm 1}College of Information Science and Electronic Engineering, Zhejiang University
}

\begin{document}
\maketitle


\begin{abstract}
Reconstructing continuous physical fields from sparse measurements is
central to scientific monitoring, inverse modeling, and digital-twin
construction.
Generative reconstruction has recently emerged as a promising paradigm
for this task by learning data-driven physical priors that complete
plausible full fields from limited observations.
However, existing methods largely assume fixed, batch conditioning,
whereas real sensing systems often produce structured streams: probes
scan local regions, instruments observe moving fields of view, and
communication constraints may leave entire frames missing.
We propose \textbf{TRACE}, a retrospective streaming generative
reconstruction framework for physical fields under structured sensing.
TRACE performs approximate Bayesian inference in a learned
continuous-coordinate latent space, converting sparse off-grid
measurements into generative latent evidence, fusing it with a
state-space temporal prior through Kalman-style filtering, and refining
under-observed past frames via retrospective smoothing.
Experiments on active matter, ocean sound-speed fields, and supernova
simulations show that TRACE matches or surpasses frame-wise generative
reconstructors, offline spatiotemporal methods, and streaming
data-assimilation baselines in reconstruction quality under temporally
sparse and spatially localized sensing protocols.
\end{abstract}


\section{Introduction}
\label{sec:intro}

Reconstructing continuous physical fields from sparse measurements is a
foundational problem in scientific machine learning, with applications
in physical monitoring, ocean sensing, astronomical simulation, active
matter, and digital-twin maintenance. The target quantity is often a
field evolving over space and time, while sensors provide only sparse,
irregular, and off-grid readings. This sparse-to-full problem is highly
ill-posed: many full fields can explain the same limited observations.
Generative reconstruction has recently become a promising paradigm for
this setting. By learning data-driven priors over plausible physical
fields, diffusion posterior sampling~\citep{chung2023dps},
diffusion-based physical-field solvers~\citep{huang2024diffusionpde},
and continuous-coordinate generative field
models~\citep{du2024confild} can recover coherent structures from
incomplete measurements, turning reconstruction from interpolation into
conditional generation under physical and observational constraints.

However, most generative reconstructors still assume a fixed, batch view
of evidence: observations are available before inference either for a
single frame or for a complete trajectory. Real sensing systems are
often different. A mobile probe scans only a local region at each time,
an imaging instrument observes a moving field of view, and power or
communication constraints may leave entire frames missing. In such
structured sensing streams, sparsity is organized over time. A single
frame is usually under-informative, and its meaning depends on evidence
observed before and after it. Existing methods do not jointly address
this regime. Frame-wise generative reconstructors provide strong field
priors but cannot propagate evidence across time. Offline
spatiotemporal reconstructors such as SDIFT~\citep{chen2025sdift}
exploit temporal context but require the full observation horizon
before reconstruction. Classical Kalman and ensemble data
assimilation~\citep{kalman1960,evensen1994sequential}, as well as
recent neural variants~\citep{rozet2023sda,xiao2024ldensf,tarumi2025dbf},
support online updates but usually rely on gridded states or explicit
dynamics rather than continuous-coordinate generative field completion
from sparse off-grid measurements.
What remains missing is a framework that keeps the learned physical
prior of generative reconstruction, accumulates weak evidence as the
stream unfolds, and revises earlier ambiguous estimates when later
measurements make them identifiable.

\begin{figure*}[t]
\centering
\includegraphics[width=0.85\textwidth]{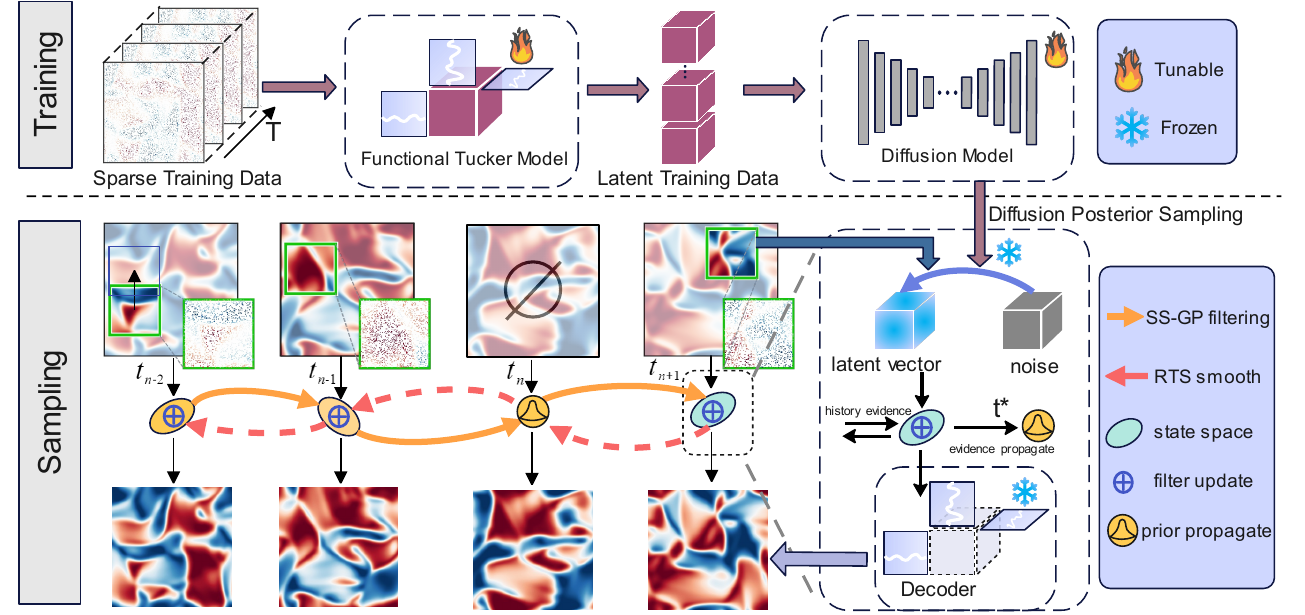}
\caption{ Overview of TRACE. Top: offline pretraining of the
continuous-coordinate field representation and latent diffusion prior.
Bottom: online streaming inference, which fuses per-frame generative
evidence with a state-space prior via Kalman filtering and refines it
via retrospective smoothing.}

\label{fig:overview}
\end{figure*}

We address this problem with \textbf{TRACE} (\textbf{T}emporal
\textbf{R}etrospective \textbf{A}ccumulation for
\textbf{C}ontinuous-field \textbf{E}stimation), a retrospective
streaming generative reconstruction framework for physical fields under
structured sensing. TRACE treats each incoming frame as latent evidence
rather than as an isolated reconstruction target. It represents fields
in a compact continuous-coordinate latent space using a functional
Tucker decoder, so sparse off-grid measurements can constrain arbitrary
spatial queries. A pretrained generative latent prior supplies plausible
single-frame completions, which are summarized as Gaussian latent
evidence through generative posterior sampling and moment matching.
TRACE then fuses this weak per-frame evidence with a Matérn
state-space temporal prior using a Kalman-style update. The forward
filter gives causal online estimates from observations seen so far.
Once later measurements arrive, a Rauch--Tung--Striebel smoothing pass
propagates information backward to refine past frames that were missing,
localized, or otherwise under-observed. TRACE thus separates causal filtering from retrospective smoothing,
 distinguishing the online estimate from the refined
offline correction. An overview is shown in Fig.~\ref{fig:overview}.

Our contributions are as follows. \textbf{(C1)} We formulate
physical-field reconstruction under \textbf{structured sensing}, where
sparse off-grid measurements arrive as temporally organized streams,
including temporally sparse frames and spatially localized moving-window
observations. \textbf{(C2)} We propose \textbf{TRACE}, a retrospective
streaming generative reconstruction framework that performs approximate
Bayesian evidence fusion in a learned continuous-coordinate latent
space. \textbf{(C3)} We develop a filtering-and-smoothing inference
mechanism that provides causal online estimates and retrospectively
refines past under-observed frames as later evidence arrives.
\textbf{(C4)} We validate TRACE on active matter, ocean sound-speed
fields, and supernova simulations, showing that TRACE matches or
surpasses frame-wise generative reconstructors, offline spatiotemporal
methods, and streaming data-assimilation baselines in reconstruction
quality under temporally sparse and spatially localized sensing
protocols.

\section{Preliminaries}
\label{sec:prelim}

\subsection{Continuous Low-Rank Tensor Representations}
\label{sec:prelim:tensor}

Many scientific fields are naturally represented as continuous
multivariate functions defined over spatial or spatiotemporal
domains. Consider a continuous $K$-variate function
$f:\mathcal{R}_1\times\cdots\times\mathcal{R}_K\to\mathbb{R}$,
where the full coordinate is written
$\mathbf{r}=(r_1,\ldots,r_K)$ with $r_k\in\mathcal{R}_k$.
A naive discretization with $I$ grid points per mode requires
$\mathcal O(I^K)$ degrees of freedom, making direct grid-based
representations prohibitively expensive as the dimensionality
increases.
Low-rank tensor representations combined with coordinate-based
neural basis functions alleviate this challenge by representing
cross-mode interactions through compact latent coefficients and
continuous coordinate decoders, as

\begin{equation}
f(r_1,\ldots,r_K)
\approx
\left\langle
\mathcal{C},
\phi^{(1)}(r_1)
\otimes
\cdots
\otimes
\phi^{(K)}(r_K)
\right\rangle,
\label{eq:continuous-tensor}
\end{equation}
where
$\phi^{(k)}(\cdot)$ denotes the coordinate-based basis function
for the $k$-th mode,
$\mathcal{C}$ is a compact coefficient tensor capturing the
multilinear interactions across modes, and
$\langle\cdot,\cdot\rangle$
denotes multilinear contraction.

This representation decouples continuous coordinate decoding from
compact latent coefficients, enabling arbitrary off-grid queries
while providing a compact latent representation suitable for
subsequent generative and temporal modeling.

\subsection{Diffusion Priors and Posterior Sampling}
\label{sec:prelim:diffusion}

Diffusion models~\cite{song2021score,karras2022elucidating}
learn a generative prior by training a denoising network to
recover clean samples from progressively corrupted observations.
Once trained, the denoiser implicitly defines a score function

\begin{equation}
s_{\boldsymbol{\theta}}(\mathbf z;\sigma)
:=
\nabla_{\mathbf z}\log p_{\sigma}(\mathbf z)
\approx
\frac{
D_{\boldsymbol{\theta}}(\mathbf z;\sigma)-\mathbf z
}{\sigma^{2}},
\label{eq:score}
\end{equation}
which estimates the gradient of the log-density of the noisy
distribution $p_{\sigma}(\mathbf z)$ and provides a learned
 unconditional generative prior for sampling.
The learned score enables unconditional generation of latent
representations. For inverse problems, however, latent generation
must additionally satisfy the available observations.
Diffusion Posterior Sampling (DPS)~\cite{chung2023dps} extends the
pretrained generative prior to conditional generation by
incorporating measurement guidance during reverse diffusion.
Given observations $\mathbf y$ generated through a forward
measurement model $G(\cdot)$, DPS approximates the posterior score
as
\begin{equation}
\nabla_{\mathbf z}
\log p(\mathbf z\mid\mathbf y;\sigma)
\approx
s_{\boldsymbol\theta}(\mathbf z;\sigma)
-
\lambda
\nabla_{\mathbf z}
\left\|
\mathbf y-
G\!\left(
D_{\boldsymbol\theta}(\mathbf z;\sigma)
\right)
\right\|_2^2,
\label{eq:dps}
\end{equation}
where $D_{\boldsymbol\theta}(\mathbf z;\sigma)$ denotes the
denoised latent estimate, $G(\cdot)$ is the forward measurement
model relating latent representations to observations, and
$\lambda>0$ controls the guidance strength. Reverse diffusion
driven by the guided score yields observation-consistent latent
samples that jointly satisfy the learned generative prior and the
measurement constraints.

\subsection{State-Space Gaussian Processes}
\label{sec:prelim:ssgp}

Gaussian processes (GPs) provide a non-parametric Bayesian prior
over continuous functions. For a temporal function $x(t)$, a
zero-mean GP is written as
$x(t)\sim\mathcal{GP}(0,\kappa(t,t'))$, where the covariance
kernel $\kappa$ specifies the correlation between function values
at different time instants. Exact GP inference, however, requires
inverting the covariance matrix and therefore scales as
$\mathcal{O}(N^{3})$ for a trajectory of length $N$.

For the widely used half-integer Mat\'ern kernel family, the GP
admits an equivalent finite-dimensional state-space
representation~\cite{hartikainen2010kalman,sarkka2019applied},
allowing it to be reformulated as a linear stochastic dynamical
system. Introducing the augmented Markov state
$\bm{\xi}(t)
=
(x(t),x^{(1)}(t),\ldots,x^{(p)}(t))^{\top}$,
which collects the function together with its first $p$
derivatives, the GP can be represented by the linear
time-invariant stochastic differential equation (LTI-SDE)

\begin{equation}
\frac{d\bm{\xi}(t)}{dt}
=
\mathbf{F}\bm{\xi}(t)
+
\mathbf{L}w(t),
\qquad
x(t)
=
\mathbf{H}\bm{\xi}(t),
\label{eq:lti-sde}
\end{equation}

where $\mathbf{F}$, $\mathbf{L}$, and $\mathbf{H}$ are determined
by the kernel hyperparameters, $w(t)$ is a scalar white-noise
process, and
$\mathbf{H}=[1,0,\ldots,0]$
projects the augmented state onto the function value.

Discretising the LTI-SDE over arbitrary timestamps
$t_1<t_2<\cdots<t_N$ yields an equivalent linear-Gaussian
Gauss--Markov process, commonly referred to as the
state-space Gaussian process (SS-GP). Writing
$\bm{\xi}_n:=\bm{\xi}(t_n)$ and
$\Delta_n:=t_n-t_{n-1}$, the state transition is

\begin{equation}
p(\bm{\xi}_n\mid\bm{\xi}_{n-1})
=
\mathcal{N}
\!\left(
\mathbf{A}_n\bm{\xi}_{n-1},\,
\mathbf{Q}_n
\right),
\label{eq:ssgp}
\end{equation}

where
$\mathbf{A}_n=\exp(\Delta_n\mathbf{F})$,
and
$\mathbf{Q}_n
=
\mathbf{P}_{\infty}
-
\mathbf{A}_n
\mathbf{P}_{\infty}
\mathbf{A}_n^{\top}$,
with
$\mathbf{P}_{\infty}$
denoting the stationary covariance obtained from the corresponding
continuous-time Lyapunov equation.

The resulting linear-Gaussian Markov process admits exact Bayesian
inference through Kalman filtering for causal online estimation
and Rauch--Tung--Striebel (RTS)
smoothing~\cite{rauch1965maximum}
for retrospective refinement, both with overall
$\mathcal{O}(N)$ computational complexity.
This state-space formulation therefore provides an efficient
temporal inference interface for streaming trajectories while
preserving the underlying Gaussian process prior.


\section{Method}
\label{sec:method}

\textbf{Problem Statement:} We consider a continuous spatiotemporal physical field
$\fieldy:\spdom\times\mathbb{R}_{+}\rightarrow\mathbb{R}$,
defined over a continuous spatial domain
$\spdom=\mathcal{R}_{1}\times\cdots\times\mathcal{R}_{K}\subset\mathbb{R}^{K}$.
Rather than observing $\fieldy$ directly, sensing produces an
irregular stream of sparse off-grid measurements arriving
sequentially at timestamps
$t_{1}<t_{2}<\cdots$.
At time $t_n$, the active sensors occupy a frame-dependent
observation region
$\obsreg\subseteq\spdom$,
producing the observation set
\begin{equation}
\Obs_n=
\bigl\{
(\rrr_{n,m},\,y_{n,m},\,t_n)
\bigr\}_{m=1}^{M_n},
\label{eq:obs-set}
\end{equation}
where
$\rrr_{n,m}\in\obsreg$
denotes an off-grid sensing location and
$y_{n,m}
=
\fieldy(\rrr_{n,m},t_n)+\epsilon_{n,m}$
is the corresponding noisy measurement.
The number of observations $M_n$ varies across frames and may
even vanish, resulting in $\Obs_n=\emptyset$ for entirely
unobserved timestamps.
The streaming history available up to time $t_n$ is therefore
$\Obshist=\{\Obs_1,\ldots,\Obs_n\}$.

Since each frame is only partially observed, reliable
reconstruction cannot be achieved independently from a single
observation set.
Instead, the objective is to continuously infer the underlying
physical field from the streaming history $\Obshist$,
producing estimates at arbitrary spatiotemporal queries
$(\rstar,\tstar)\in\spdom\times\mathbb{R}_{+}$ while allowing
newly arrived observations to retrospectively refine previously
under-observed frames.

\subsection{Streaming Reconstruction as Latent Inference}
\label{sec:method:latent}

Performing sequential inference directly in the original physical
field space is challenging under streaming sparse sensing:
irregular off-grid observations vary across frames, preventing a
common spatial discretisation for temporal aggregation;
reconstructing a continuous field from an individual under-observed
frame is ill-posed without exploiting intrinsic spatial structure;
and recursive inference over the original high-dimensional
continuous function space is computationally prohibitive.

To address these challenges, TRACE first reformulates streaming
field reconstruction as inference over a compact latent trajectory
using the Functional Tucker Model (FTM)~\cite{fang2024funbat},
which serves as a low-rank model, encoding each observation frame
into a fixed-dimensional latent space.

Specifically, for a physical field
$\fieldy$ defined over the $K$-mode spatial domain
$\spdom=\mathcal{R}_1\times\cdots\times\mathcal{R}_K$
with preset multilinear ranks
$\{R_k\}_{k=1}^{K}$,
FTM parameterises each frame $t_n$ by a frame-specific Tucker core
$\Wcore_n\in\mathbb{R}^{R_1\times\cdots\times R_K}$ together with
shared continuous basis functions
$\boldsymbol{\phi}^{(k)}_{\theta_k}:
\mathcal{R}_k\rightarrow\mathbb{R}^{R_k}$.
The field value at an arbitrary continuous coordinate
$\rrr=(r_1,\ldots,r_K)$ is represented as

\begin{equation}
\fieldy(\rrr,t_n)
\approx
\mathrm{vec}(\Wcore_n)^\top
\left(
\boldsymbol{\phi}^{(1)}_{\theta_1}(r_1)
\otimes
\cdots
\otimes
\boldsymbol{\phi}^{(K)}_{\theta_K}(r_K)
\right),
\label{eq:ftm-entry}
\end{equation}

where $\mathrm{vec}(\cdot)$ denotes vectorisation and
$\otimes$ denotes the Kronecker product.
Unlike tensor decompositions defined on fixed spatial grids, the
continuous basis functions can be evaluated at arbitrary query
coordinates, naturally accommodating different off-grid sensing
layouts while preserving a unified latent representation.

The FTM is pretrained on the training data
$\Datacorpus=\{\Obs^{(b)}\}_{b=1}^{B}$,
where each trajectory
$\Obs^{(b)}=\{\Obs_n^{(b)}\}_{n=1}^{N^{(b)}}$
contains the complete observation sequence.
The basis functions
$\{\boldsymbol{\phi}^{(k)}_{\theta_k}\}_{k=1}^{K}$
and Tucker cores
$\{\Wcore_n^{(b)}\}_{b,n}$
are jointly learned by reconstructing the underlying physical
fields from $\Datacorpus$. The detailed optimisation procedure and network architectures are
deferred to Appendix~\ref{app:impl:ftm}.

To obtain a unified latent vector for subsequent inference objectives,
we vectorise each Tucker core as 
$\zlat_n=\mathrm{vec}(\Wcore_n)\in\mathbb{R}^{d}$,
where $d=\prod_{k=1}^{K}R_k$, yielding the latent training dataset
$\mathcal{Z}=\{\zlat_n^{(b)}\}_{b,n}$.
The physical field is recovered through the decoder
\begin{equation}
\hat{\fieldy}(\rstar,t_n)
=
\Gobs(\rstar)^{\top}\zlat_n,
\label{eq:decoder}
\end{equation}
where
$\Gobs(\rstar)
=
\bigotimes_{k=1}^{K}
\boldsymbol{\phi}_{\theta_k}^{(k)}(r_k^{\star})
\in\mathbb{R}^{d}$
evaluates the frozen basis functions at the query coordinate
$\rstar=(r_1^{\star},\ldots,r_K^{\star})$.
This decoder provides a unified interface for both sparse
measurements and continuous queries through the same operator
$\Gobs(\cdot)$.

Consequently, streaming field reconstruction reduces to inferring
the latent trajectory $\{\zlat_n\}_{n=1}^{N}$ from the incoming
observation stream. We formulate this as two complementary
Bayesian inference objectives.
The \emph{causal filtering} objective estimates the current latent
vector using only the observations available up to the current
timestamp,
\begin{equation}
q_n^{\mathrm{filt}}(\zlat_n)
\;\approx\;
p(\zlat_n\mid\Obshist),
\label{eq:obj-filter}
\end{equation}
where $\Obshist=\{\Obs_1,\ldots,\Obs_n\}$ denotes the streaming
history observed so far.
To further improve previously estimated accuracy, the
\emph{retrospective smoothing} objective incorporates future
observations,
\begin{equation}
q_k^{\mathrm{smooth}}(\zlat_k)
\;\approx\;
p(\zlat_k\mid\Obs_{1:N}),
\label{eq:obj-smooth}
\end{equation}
where  $k=1,\ldots,N $ and $\Obs_{1:N}$ denotes all observations received up to the
current stream horizon.

TRACE achieves these two objectives through a unified latent
inference pipeline. Each incoming observation set is first
converted into probabilistic latent evidence by a pretrained
generative prior. The latent evidence is then recursively fused
with a state-space temporal prior through forward filtering, while
backward smoothing retrospectively propagates future information to
refine previously estimated latent states. The inferred latent
trajectory is finally decoded into continuous physical fields via
Eq.~\eqref{eq:decoder} whenever observations are assimilated or
field values are queried.

\subsection{Generative Latent Evidence}
\label{sec:method:dps}

When a new sparse off-grid observation frame $\Obs_n$ arrives,
our goal is to represent it individually in latent space as probabilistic
evidence.
Given the pretrained
decoder of Eq.~\eqref{eq:decoder}, each observation frame induces
a linear measurement operator connecting the latent vector to the
observed values:
\begin{equation}
\mathbf{y}_n = \Gobs_n\zlat_n + \boldsymbol{\epsilon}_n,
\qquad
\boldsymbol{\epsilon}_n\sim\mathcal{N}(\mathbf{0},
\sigma_{\text{obs}}^{2}\mathbf{I}_{M_n}),
\label{eq:meas-model}
\end{equation}
where $\Gobs_n\in\mathbb{R}^{M_n\times d}$ is the matrix whose
$m$-th row is $\Gobs(\rrr_{n,m})^{\top}$, and
$\mathbf{y}_n=(y_{n,1},\ldots,y_{n,M_n})^{\top}$.

However, when individual frames are under-informative, covering only
a limited localised region, the posterior $p(\zlat_n\mid\Obs_n)$ is
ill-posed without a strong prior over the latent space. We employ a pretrained unconditional
latent diffusion model trained offline on $\mathcal{Z}$ following
EDM~\cite{karras2022elucidating} (see Appendix~\ref{app:impl:edm});
the resulting denoiser $D_{\boldsymbol{\theta}}$ defines a score-based
prior $s_{\boldsymbol{\theta}}(\zlat;\sigma)=\nabla_{\zlat}\log p(\zlat;\sigma)$
that is frozen during streaming inference.
To sample from this posterior given the current observation, we use
Diffusion Posterior Sampling (DPS)~\cite{chung2023dps}, which
augments the prior score with a likelihood-guidance term derived
from the measurement model of Eq.~\eqref{eq:meas-model}. Following
the formulation of Eq.~\eqref{eq:dps} with measurement operator
$\Gobs_n$, the conditional score is approximated as
\begin{multline}
\nabla_{\zlat_n}\!\log p(\zlat_n\mid\mathbf{y}_n;\sigma)
\;\approx\;
s_{\boldsymbol{\theta}}(\zlat_n;\sigma)
\\
-\;\zeta\,
\nabla_{\zlat_n}\bigl\|\mathbf{y}_n
-\Gobs_n\,
\hat{\zlat}_{0}(\zlat_n;\sigma)\bigr\|_{2}^{2},
\label{eq:dps-trace}
\end{multline}
where
$\hat{\zlat}_{0}(\zlat_n;\sigma)=D_{\boldsymbol{\theta}}(\zlat_n;\sigma)$
is the Tweedie estimate of the clean latent and $\zeta>0$ sets the
guidance strength.  Running reverse diffusion with this guided score for $S$
independent noise initialisations, which are computed in parallel along the
batch dimension, yields observation-consistent latent samples
$\{\zlat_n^{(s)}\}_{s=1}^{S}$ from this approximate posterior.

The $S$ DPS samples define an empirical distribution over the latent
space. To obtain a compact probabilistic representation, we fit a
Gaussian through moment matching:
\begin{equation}
q_n^{\mathrm{gen}}(\zlat_n\mid\Obs_n)
=
\mathcal{N}\bigl(
\hat{\boldsymbol{\mu}}_n^{\mathrm{gen}},\,
\hat{\boldsymbol{\Sigma}}_n^{\mathrm{gen}}
\bigr),
\label{eq:moment-match}
\end{equation}
where
$\hat{\boldsymbol{\mu}}_n^{\mathrm{gen}}=\frac{1}{S}\sum_{s}\zlat_n^{(s)}$
and
$\hat{\boldsymbol{\Sigma}}_n^{\mathrm{gen}}=\frac{1}{S-1}\sum_{s}(\zlat_n^{(s)}-\hat{\boldsymbol{\mu}}_n^{\mathrm{gen}})(\zlat_n^{(s)}-\hat{\boldsymbol{\mu}}_n^{\mathrm{gen}})^{\top}$.
We term
$q_n^{\mathrm{gen}}$
the per-frame \emph{generative latent evidence},
a Gaussian summary of the posterior information conveyed by the
current observation frame $\Obs_n$ under the learned field prior.


\subsection{Forward Online Filtering}
\label{sec:method:filter}

Under structured streaming sensing, observations may arrive irregularly
or become entirely missing over extended intervals, leaving the
evidence unavailable at some timestamps. Moreover, the per-frame
evidence $q_n^{\mathrm{gen}}$ is generated independently for each
frame and does not propagate information across time. To obtain a
causal filtering posterior $q_n^{\mathrm{filt}}(\zlat_n)\approx
p(\zlat_n\mid\Obs_{1:n})$ that accumulates information as the stream
unfolds, TRACE combines the current generative evidence with a temporal
belief propagated from previous frames.

To this end, we place an independent Mat\'ern-$3/2$ state-space
Gaussian process  prior on each latent dimension
$j=1,\ldots,d$ with shared hyperparameters $(\sigma_f^2,\ell)$.
Following the LTI-SDE equivalence of the stationary Mat\'ern kernel, each scalar latent trajectory
$z^{(j)}(t)$ is represented by the augmented state
$\boldsymbol{\xi}^{(j)}(t)
=(z^{(j)}(t),\dot z^{(j)}(t))^{\top}$.
The resulting discrete Gauss--Markov transition propagates the posterior parameters through the recursion
$\boldsymbol{\xi}_{n}^{(j)}\mid\boldsymbol{\xi}_{n-1}^{(j)}\sim\N(\mathbf{A}_{n}\boldsymbol{\xi}_{n-1}^{(j)},\mathbf{Q}_{n})$.
Since the transition matrices are determined by the interval
$\Delta_n=t_n-t_{n-1}$, the same temporal prior naturally handles
irregular streaming arrivals without retraining. 

Stacking the $d$ latent dimensions gives the predicted block-diagonal
augmented-state distribution $(\mathbf m_n^-,\mathbf P_n^-)$ from the
previous filtering state $(\mathbf m_{n-1},\mathbf P_{n-1})$.
Projecting through $\mathbf H=[1,0]$ yields the latent-space predicted
prior:
\begin{equation}
p_n^-(\zlat_n)
=
\mathcal{N}
(\boldsymbol{\mu}_n^-,
\boldsymbol{\Sigma}_n^-),
\label{eq:pred-prior}
\end{equation}
which encodes the temporal belief before assimilating the current frame.

We then combine the predicted prior $p_n^{-}$ and the current observation 
generative evidence $q_n^{\mathrm{gen}}$ via a tempered
product-of-experts rule:
\begin{equation}
q_{n}^{\mathrm{filt}}(\zlat_{n})
\;\propto\;
\bigl[p_{n}^{-}(\zlat_{n})\bigr]^{\alpha}\,
\bigl[q_{n}^{\mathrm{gen}}(\zlat_{n}\mid\Obs_{n})\bigr]^{\beta},
\label{eq:fusion}
\end{equation}
where $\alpha,\beta>0$ control the relative weight assigned to the
temporal prior and the generative evidence.
Since both factors are
Gaussian, the fused posterior remains Gaussian,
$q_n^{\mathrm{filt}}=\mathcal{N}(\boldsymbol{\mu}_n,\boldsymbol{\Sigma}_n)$,
with precision
\begin{equation}
\boldsymbol{\Sigma}_{n}^{-1}
= \alpha\bigl(\boldsymbol{\Sigma}_{n}^{-}\bigr)^{-1}
+ \beta\bigl(\hat{\boldsymbol{\Sigma}}_{n}^{\mathrm{gen}}\bigr)^{-1},
\label{eq:fusion-precision}
\end{equation}
and mean $\boldsymbol{\mu}_n$ providing the fused latent estimate,
which can be decoded to the physical field.
When $\alpha=\beta=1$, Eq.~\eqref{eq:fusion} reduces to the
standard Gaussian product of experts, algebraically equivalent to
an information-form Kalman update with $q_n^{\mathrm{gen}}$ as a
pseudo-observation (proof in Appendix~\ref{app:kf-proof}).

To continue the temporal recursion, the fused latent posterior is
lifted back to the augmented state through a per-dimension Kalman
measurement update, treating the $j$-th component
$(\mu_{n,j},\Sigma_{n,jj})$ of $q_{n}^{\mathrm{filt}}$ as a virtual
measurement of the value component $z_{n}^{(j)}=\mathbf{H}\boldsymbol{\xi}_{n}^{(j)}$:
\begin{equation}
(\mathbf{m}_{n},\mathbf{P}_{n})
= \mathrm{KalmanUpdate}\bigl(
\mathbf{m}_{n}^{-},\mathbf{P}_{n}^{-};\,
\boldsymbol{\mu}_{n},\,\boldsymbol{\Sigma}_{n}
\bigr),
\label{eq:kalman-abstract}
\end{equation}
implemented as $d$ parallel two-dimensional Kalman recursions
(explicit per-dimension formulae in Appendix~\ref{app:kf-proof}).
The resulting state summarizes all information available up to $t_n$,
completing one cycle of the forward online filtering recursion,
which is repeated as the start point for the next incoming frame without accessing future observations

When $M_n=0$ for a streaming frame, no generative evidence is
available ($q_n^{\mathrm{gen}}$ is undefined). TRACE therefore skips
the fusion step and directly sets
$q_n^{\mathrm{filt}}=p_n^{-}$,
using the temporal prediction as the best causal estimate.
These under-observed frames remain associated with the propagated
latent trajectory and can later be refined by the retrospective
smoother.

\subsection{Retrospective Refinement by Backward Smoothing}
\label{sec:method:smooth}

The forward filter provides a strictly causal estimate using only
the observation prefix $\Obs_{1:n}$. However, causality prevents it
from exploiting future measurements that may reveal latent
structures hidden in earlier uncertain states, particularly when
frames are under-informative or entirely missing. Given an available observation horizon $N$, TRACE therefore performs
retrospective refinement by propagating information from later
observations backward through the latent state-space dynamics, yielding the smoothing posterior
$q_{k}^{\mathrm{smooth}}(\zlat_{k}) \approx p(\zlat_{k}\mid\Obs_{1:N})$
for $k=1,\ldots,N$.
Because the temporal prior is linear-Gaussian
in the augmented state, this posterior is obtained exactly by the
standard Rauch--Tung--Striebel (RTS)~\cite{rauch1965maximum}
recursion (explicit update formulae in Appendix~\ref{app:rts});
this backward recursion is not a separate heuristic, but rather the
inference mechanism for propagating future evidence into past estimates.

Once the smoothed augmented trajectory is available, the same Markov
structure supports prediction at any continuous query time
$t^{\star}\in(t_{1},t_{N})$ through an SS-GP bridge between the
nearest smoothed neighbours. The resulting predictive distribution
is Gaussian, and composing its mean with the decoder yields the
field estimate
\begin{equation}
\hat{\fieldy}(\rstar,t^{\star})
= \Gobs(\rstar)^{\top}\,\mathbb{E}[\zlat(t^{\star})],
\label{eq:field-query}
\end{equation}
for any continuous spatial query $\rstar$ and any time $t^{\star}$
covered by the observation span. The bridge construction is
provided in Appendix~\ref{app:bridge}.

\begin{table*}[t]
\centering
\small
\caption{Field-domain RMSE under structured sensing ($\rho=1\%$).
\textbf{(a)} Temporally sparse: Miss (gap $n{=}3$, one active frame in
four) and Blk-$L$ (contiguous blackout of $L\in\{5,10\}$ frames).
\textbf{(b)} Spatially localized: moving window at local density
$\rho_{\text{loc}}=15\%$, with S-curve single-pass and circular
sweeps of one and three loops.  Lower is better; \textbf{bold} is
best per column; ``---'' marks cells a per-frame method cannot
reconstruct.}
\label{tab:structured}
\begin{tabular}{l ccc ccc ccc}
\toprule
& \multicolumn{3}{c}{Active Matter} & \multicolumn{3}{c}{Ocean} & \multicolumn{3}{c}{Supernova}\\
\cmidrule(lr){2-4}\cmidrule(lr){5-7}\cmidrule(lr){8-10}

\multicolumn{10}{c}{\textbf{(a) Temporally Sparse Observations}}\\[-2pt]
\cmidrule(lr){1-10}
& Miss & Blk-5 & Blk-10 & Miss & Blk-5 & Blk-10 & Miss & Blk-5 & Blk-10\\
\midrule
LRTFR & 0.413 & 0.258 & 0.461 & 0.205 & 0.163 & 0.177 & 0.561 & 0.430 & 0.503\\
MMGN & 0.313 & 0.217 & 0.360 & 0.097 & 0.080 & \textbf{0.092} & \textbf{0.318} & 0.318 & \textbf{0.320}\\
SDIFT & 0.490 & 0.391 & 0.530 & 0.158 & 0.127 & 0.145 & 0.436 & 0.398 & 0.424\\
DBF   & 0.960 & 0.814 & 0.873 & 0.809 & 0.627 & 0.709 & 0.855 & 0.809 & 0.845\\
\midrule
TRACE-Frame   & --- & --- & --- & --- & --- & --- & --- & --- & ---\\
TRACE-Filter  & 0.429 & 0.211 & 0.385 & 0.191 & 0.108 & 0.257 & 0.389 & 0.375 & 0.539\\
TRACE-Smoother & \textbf{0.290} & \textbf{0.186} & \textbf{0.332} & \textbf{0.095} & \textbf{0.059} & 0.124 & 0.334 & \textbf{0.312} & 0.358\\

\midrule
\multicolumn{10}{c}{\textbf{(b) Spatially Localized Observations}}\\[-2pt]
\cmidrule(lr){1-10}
& S-curve & 1-loop & 3-loop & S-curve & 1-loop & 3-loop & S-curve & 1-loop & 3-loop\\
\midrule
LRTFR & 0.769 & 0.807 & 0.693 & 0.245 & 0.351 & 0.153 & 0.917 & 0.929 & 0.785\\
MMGN  & 0.767 & 0.782 & 0.777 & 0.285 & 0.277 & 0.271 & \textbf{0.321} & \textbf{0.321} & 0.321\\
SDIFT & 0.851 & 0.793 & 0.655 & 0.281 & 0.275 & 0.290 & 0.430 & 0.430 & 0.432\\
DBF   & 0.962 & 0.956 & 0.950 & 0.440 & 0.461 & 0.439 & 0.734 & 0.729 & 0.724\\
\midrule
TRACE-Frame   & 0.580 & 0.581 & 0.577 & 0.114 & 0.147 & 0.141 & 0.429 & 0.419 & 0.414\\
TRACE-Filter  & 0.530 & 0.564 & 0.510 & 0.120 & 0.125 & 0.112 & 0.371 & 0.352 & 0.342\\
TRACE-Smoother & \textbf{0.509} & \textbf{0.537} & \textbf{0.476} & \textbf{0.097} & \textbf{0.106} & \textbf{0.088} & 0.328 & 0.334 & \textbf{0.318}\\
\bottomrule
\end{tabular}
\end{table*}



\smallskip
\noindent\textbf{Computational cost.}
With diagonal Gaussian evidence summaries, the SS-GP filtering update
costs $\mathcal{O}(d)$ per frame, while the RTS backward pass costs
$\mathcal{O}(dN)$ over a stream of length $N$, excluding the parallel
generative sampling cost. The full-covariance formulation, diagonal
approximation, and per-component runtime are detailed in
Appendix~\ref{app:diag-bias}.


\section{Related Work}
\label{sec:related}

\paragraph{Functional Tensor Decomposition.}
Functional generalisations of
Tucker~\citep{fang2022bctt,fang2023sftl} and tensor-train  
decomposition lift discrete factor matrices to continuous-coordinate
basis functions, enabling reconstruction at arbitrary off-grid
queries~\citep{luo2024lrtfr,fang2024funbat,chen2025gret}. Implicit  
neural representations such as MMGN~\citep{luo2024mmgn} follow the  
same separation-of-variables structure with multiplicative neural
modulations. These methods fit per-trajectory latents against batch
observations and lack a learned generative prior over the
field-trajectory ensemble.

\begin{figure}[t]
\centering
\includegraphics[width=\linewidth]{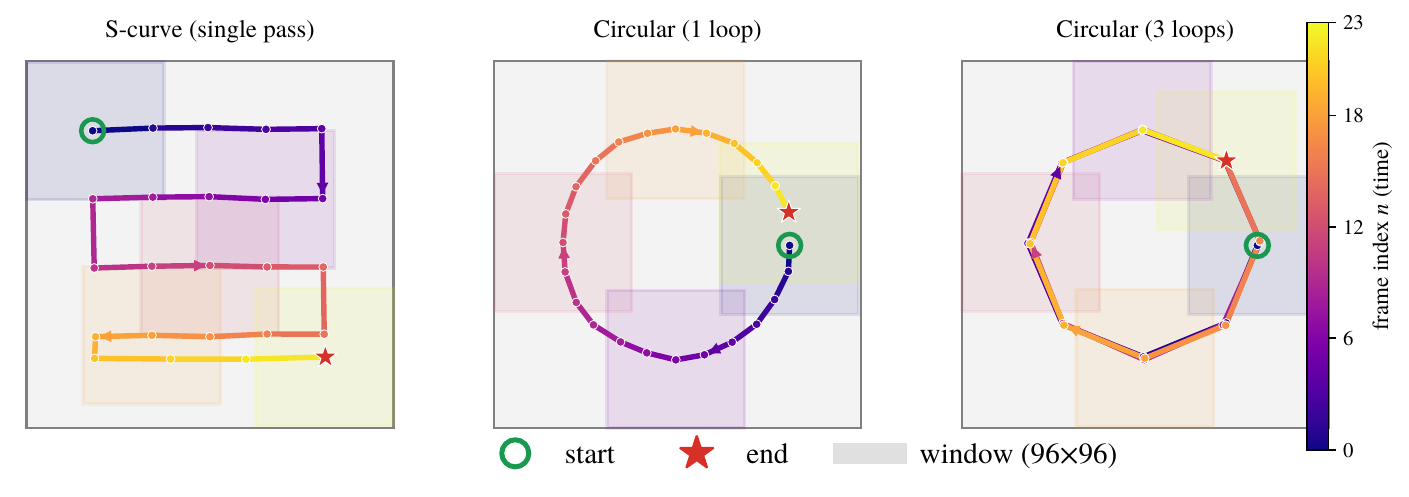}
\caption{Moving-window trajectories on Active Matter 
($256\times256$ domain, $96\times96$ window). 
The centre path (colour-coded by time) and 
shaded squares mark representative window positions;
 circular revisits accumulate coverage over time.}

\label{fig:window_traj}
\end{figure}

\paragraph{Generative Reconstruction of Physical Fields.}
Diffusion posterior sampling~\citep{chung2023dps} couples a pretrained  
score model with likelihood gradients to steer reverse sampling toward
observation-consistent fields, advancing sparse-to-dense
reconstruction~\citep{huang2024diffusionpde,li2024s3gm,long2025acmfd}.  
Neural-field latent diffusion further conditions generation on
instantaneous sparse measurements to synthesize coherent
spatiotemporal turbulence~\citep{du2024confild}.  
Most closely, SDIFT~\citep{chen2025sdift} pairs a functional Tucker  
latent with a diffusion prior and enforces inter-frame consistency
through batch Gaussian-process regression with message-passing
posterior sampling. These approaches are inherently offline,
presupposing the full observation horizon at inference.

\paragraph{Streaming and Retrospective Inference.}
Classical data assimilation combines an explicit dynamical model with
sparse observations via Kalman
recursions~\citep{kalman1960,evensen1994sequential}, enabling  
sequential state estimation of evolving physical fields. Recent work
embeds learned priors into Bayesian filtering through score
models~\citep{rozet2023sda,bao2024ensf,si2025latentensf,xiao2024ldensf}  
or deep state-space filters~\citep{tarumi2025dbf}; most of these  
methods, however, assume gridded measurements and run strictly causal
forward filtering, with no retrospective refinement of past estimates.
Two lines of work instead offer retrospective refinement. On the
generative side, building on per-frame-noise sequence
diffusion~\citep{chen2024diffusionforcing},  
ForcingDAS~\citep{jia2026forcingdas} learns a joint-trajectory  
diffusion prior whose noise schedule selects filtering, fixed-lag, or
full-sequence smoothing at inference, albeit with causal-only smoothing
on gridded pixel fields. On the Gaussian side, the state-space GP
equivalence~\citep{hartikainen2010kalman} enables exact forward  
filtering and backward smoothing in linear time and underlies online
tensor imputation~\citep{fang2024bayotide}; yet these Gaussian methods  
operate in the raw data domain and lack a learned generative prior over
field trajectories.

TRACE bridges these research directions by combining continuous
off-grid field representations, latent generative priors, and
retrospective temporal inference for streaming reconstruction.



\section{Experiments}

\subsection{Experimental Setup}
\label{sec:exp:setup}

\begin{figure*}[t]
\centering
\includegraphics[width=0.75\textwidth]{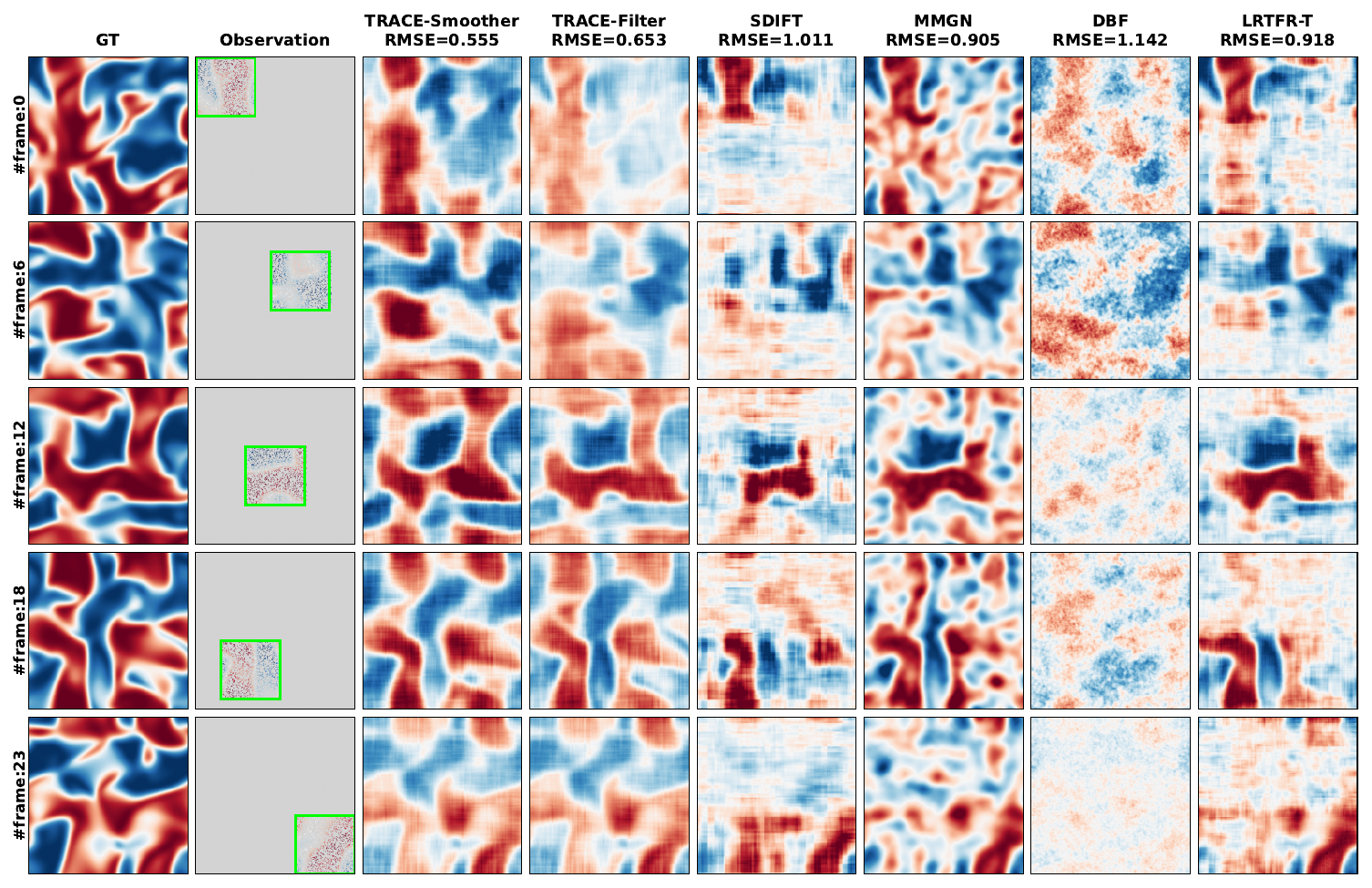}
\caption{Qualitative reconstruction on Active Matter under a one-lap circular moving window.}
\label{fig:qual_am}
\end{figure*}


\begin{figure*}[t]
\centering
\scalebox{0.80}{  
  \begin{minipage}{0.33\textwidth}\centering
    \includegraphics[width=\linewidth]{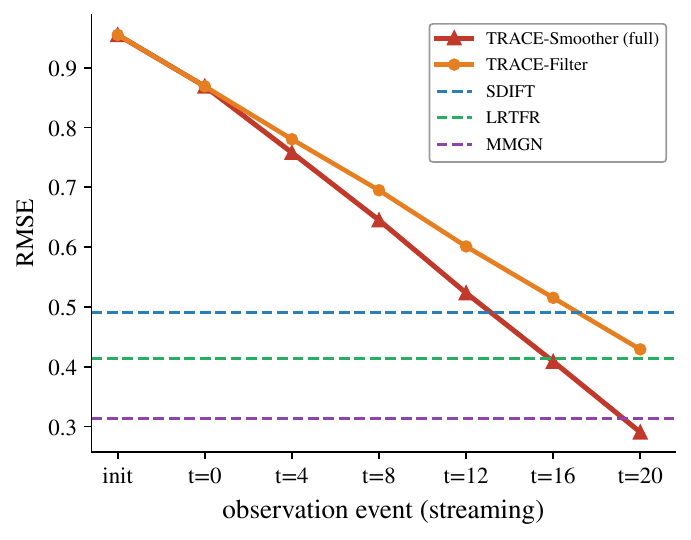}\\[2pt]
    {\small (a) Online accumulation}
  \end{minipage}\hfill
  \begin{minipage}{0.33\textwidth}\centering
    \includegraphics[width=\linewidth]{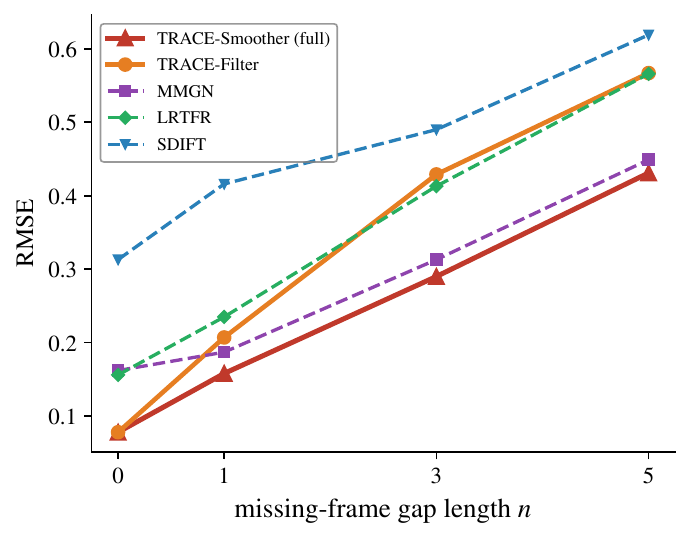}\\[2pt]
    {\small (b) Missing-frame gaps }
  \end{minipage}\hfill
  \begin{minipage}{0.33\textwidth}\centering
    \includegraphics[width=\linewidth]{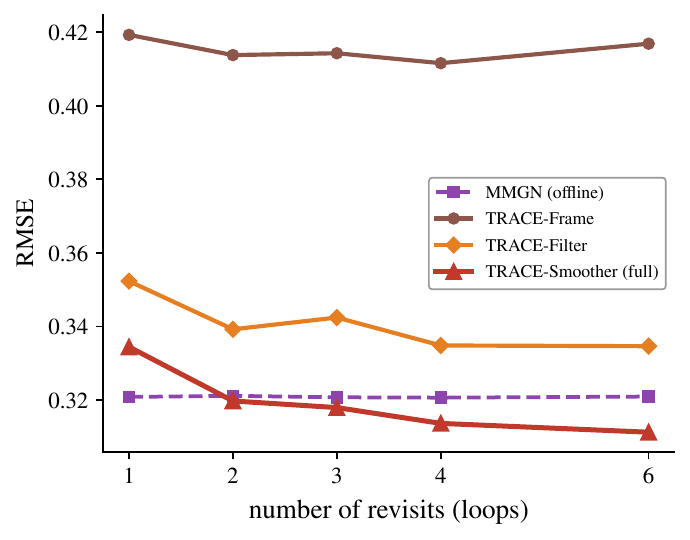}\\[2pt]
    {\small (c) Revisit frequency }
  \end{minipage}
}
\caption{Regime-structured analysis of TRACE.}
\label{fig:regime}
\end{figure*}


\paragraph{Datasets.}
We evaluate TRACE on three spatiotemporal physical-field benchmarks spanning distinct physical systems, temporal dynamics, and spatial dimensionalities.
\textbf{Active Matter} contains simulated rod-like particle dynamics in a Stokes fluid ($256\times256$, $T=24$), representing rapidly evolving laboratory-scale phenomena.
\textbf{Ocean} consists of Pacific sound-speed reanalysis fields ($5\times38\times76$, $T=24$), exhibiting moderately varying large-scale ocean dynamics.
\textbf{Supernova} contains astrophysical blast-wave simulations ($64^3$, $T=16$), whose temporal evolution is comparatively slow.
Together, these datasets cover both 2D and 3D physical fields and range from highly dynamic to nearly stationary regimes.
Dataset preprocessing, train/test splits, and latent tensor ranks are provided in Appendix~\ref{app:datasets}.

\paragraph{Structured sensing protocols.}
We evaluate TRACE under two structured sensing protocols that examine
its ability to accumulate information across time.
\textbf{Temporally sparse observations:}
only a subset of frames receives measurements, while the remaining
frames are entirely unobserved.
This protocol evaluates whether temporal evidence can be propagated
across missing intervals and whether later observations can
retrospectively refine earlier estimates.
\textbf{Spatially localized observations:}
each frame observes only a moving local region of the spatial domain.
Since no individual frame provides global coverage, accurate
reconstruction requires progressively integrating partial spatial
evidence as observations arrive.
We detail the observation patterns and parameter settings in Sections~\ref{sec:exp:temporal} and~\ref{sec:exp:spatial}, alongside the corresponding results.

\paragraph{Baselines.}
Existing sparse physical-field reconstruction methods mainly operate in
an offline setting, where inference has access to the complete
observation horizon.
We compare against representative offline reconstruction methods from
three paradigms:
(i) \textbf{LRTFR}~\cite{luo2024lrtfr}, a low-rank functional Tucker
reconstruction method with temporal total-variation regularization;
(ii) \textbf{MMGN}~\cite{luo2024mmgn}, an implicit neural reconstruction
method based on modulated neural fields; and
(iii) \textbf{SDIFT}~\cite{chen2025sdift}, a generative spatiotemporal
reconstruction method combining a functional Tucker latent
representation with GP-correlated latent diffusion and posterior
sampling.
For sequential inference, we additionally compare with the closest
available streaming baseline,
\textbf{DBF}~\cite{tarumi2025dbf}, a deep Bayesian filtering method for
state estimation from streaming observations.
For controlled comparison, LRTFR and SDIFT share the same pretrained FTM
basis as TRACE.
Additional implementation details are provided in
Appendix~\ref{app:baselines}.

\paragraph{TRACE variants and evaluation.}
We evaluate three variants to isolate the contribution of each temporal
inference component.
\textbf{TRACE-Frame} performs independent per-frame generative
reconstruction without temporal inference.
\textbf{TRACE-Filter} augments it with the strictly causal SS-GP forward
filter, enabling online temporal evidence accumulation.
\textbf{TRACE-Smoother} further applies RTS backward smoothing to
retrospectively refine earlier estimates using later observations.
Accordingly, the improvement from TRACE-Frame to TRACE-Filter measures
the benefit of causal temporal accumulation, while the improvement from
TRACE-Filter to TRACE-Smoother quantifies the value of retrospective
refinement.
We report field-domain RMSE averaged over all frames and held-out test
trajectories as the primary evaluation metric.

\subsection{Temporally Sparse Observations}
\label{sec:exp:temporal}

We instantiate temporally sparse sensing through two complementary
patterns: \emph{Miss}, where a repeating cycle of one observed frame
followed by $n{=}3$ missing frames leaves every fourth frame active;
and \emph{Blk-$L$}, a single contiguous blackout of $L\in\{5,10\}$
frames in an otherwise fully-observed stream.  Active frames receive
sparse off-grid observations at density $\rho=1\%$ of the full
spatial domain.  Panel (a) of Table~\ref{tab:structured} reports
field-domain RMSE.

The results show that since TRACE-Frame is not applicable on
unobserved frames, temporal inference becomes the deciding factor.
TRACE-Filter produces valid online estimates by propagating latent
states through the SS-GP forward recursion. However, because
inference remains strictly causal, the SS-GP prior decays over
extended missing intervals, and accuracy deteriorates on longer gaps
and on rapidly evolving dynamics such as Active Matter.
TRACE-Smoother consistently outperforms both TRACE-Frame and
TRACE-Filter, with the RTS backward pass retrospectively refining
earlier latent states by integrating observations arriving after each
missing interval. This gain is most pronounced on Active Matter, where
TRACE-Smoother matches or surpasses the strongest offline baseline
across all Miss and Blk-$L$ settings. On the slower-varying Ocean 
and the volumetric Supernova, offline methods such as MMGN remain 
competitive under the longest blackouts, as their latent-space
interpolation is smoother when adjacent frames evolve slowly and 
the full observation horizon is available.
Under rapidly evolving dynamics, however, such interpolation fails, and TRACE-Smoother
achieves the largest gains.

Qualitative examples under temporal sparsity are provided in
Appendix~\ref{app:exp:qual}.

\subsection{Spatially Localized Observations}
\label{sec:exp:spatial}

The second structured regime keeps every frame active but restricts the
observation region to a moving local window that translates over time;
no single frame covers the global domain, so a coherent field must be
accumulated as the window sweeps through space.  Each frame observes the
window interior at local density $\rho_{\text{loc}}=15\%$.  Window sizes
are dataset-specific: $96{\times}96$ on Active Matter, a window
spanning the full 5-layer depth on Ocean, and $32^3$ on Supernova.
Panel (b) of Table~\ref{tab:structured} reports field-domain RMSE for
three trajectories (S-curve single-pass, circular sweeps of one and
three loops).
 
The results show that TRACE-Frame already surpasses the strongest
offline methods on Active Matter and Ocean, indicating the strength of
the generative latent prior. TRACE-Filter improves upon TRACE-Frame as
evidence accumulates over the streaming observations, and TRACE-Smoother
further improves upon TRACE-Filter through retrospective refinement. On
the 3D Supernova, the offline method MMGN remains competitive for the
single-pass S-curve and one-loop trajectories, as its full-trajectory
optimization benefits from the near-static high-dimensional structure.
However, as the number of revisits increases to three loops,
TRACE-Smoother surpasses MMGN, demonstrating that accumulated coverage
eventually compensates for limited per-frame observability. Notably, more
frequent revisits provide richer temporal evidence and lower
reconstruction error, consistent with the information-accumulation
mechanism of our framework.  Figure~\ref{fig:qual_am} illustrates this
accumulation qualitatively on a circular sweep.

\subsection{Regime-Structured Analysis}
Finally, we analyse how the structure of the observation process itself
shapes performance (Fig.~\ref{fig:regime}).\textbf{(a) Online accumulation} (Active Matter, $\rho{=}1\%$, Miss $n{=}3$):
as the stream unfolds,
TRACE-Smoother progressively refines its full-trajectory estimate
using all observations accumulated so far, surpassing offline batch
methods before the complete horizon is observed.
 \textbf{(b) Missing-frame gaps} (Active Matter, $\rho=1\%$): as the gap
length increases, TRACE-Filter degrades significantly, while TRACE-Smoother
retrospectively refines earlier frames from later evidence, with the
performance gap over the causal filter widening as the temporal structure
becomes more severe. \textbf{(c) Revisit frequency} (Supernova, moving window,
$\rho_{\text{loc}}{=}15\%$): we compare the effect of revisit speed
on reconstruction. The
strongest offline baseline (MMGN) remains nearly flat with respect to
revisit frequency, whereas TRACE-Smoother improves as revisits become more
frequent, because the SS-GP prior accumulates information across
revisits to fill in unobserved regions.

\subsection{Additional Results}
We provide further analyses in Appendix~\ref{app:extra}. The appendix
includes: (i) dense-time random sparse control results, confirming that
TRACE remains competitive when every frame is already informative;
(ii) robustness to varying observation sparsity and sensor noise;
(iii) ablations over key hyperparameters ($\alpha$, $\ell$, $S$);
(iv) runtime analysis; and (v) additional
qualitative examples across all three datasets.

\section{Conclusion}
\label{sec:conclusion}

We introduced \textbf{TRACE}, a streaming generative framework
combining continuous-coordinate latent decoding, generative posterior
sampling, SS-GP temporal fusion, and RTS smoothing for physical fields
under structured sensing. Experiments on active matter, ocean, and
supernova benchmarks show TRACE achieves the strongest overall
performance, with largest gains under rapidly evolving dynamics and
repeated spatial coverage while remaining competitive in slowly varying
regimes. Future work includes active moving-window strategies that
steer the sensor toward informative regions rather than following
predefined trajectories, and replacing the non-parametric SS-GP prior
with a learned physics-informed neural dynamics model that predicts
latent-space field evolution from sparse observations.

\bibliography{references}

\clearpage
\appendix


\section{Mathematical Derivations}
\label{app:derivations}

This appendix collects the derivations supporting the
state-space inference machinery of
Sec.~\ref{sec:method:filter}.
We first recall, in standard spectral form, the
Hartikainen--Särkkä equivalence that turns a Matérn-$\nu$
Gaussian-process prior into an equivalent finite-dimensional
LTI-SDE (\S\ref{app:ssgp-spectral}). We then specialise to
$\nu=3/2$ and write out the closed-form matrices used throughout
this paper (\S\ref{app:matern32-closed-form}).
The remaining three subsections analyse the tempered fusion
introduced in Sec.~\ref{sec:method:filter}: its closed form and its
Kalman corollary (\S\ref{app:kf-proof}), the rationale for the
diagonal-fusion approximation (\S\ref{app:diag-bias}), and finally
the algorithmic components for backward smoothing and off-grid
querying (\S\ref{app:rts-bridge}).
Throughout, we use the same symbols as Sec.~\ref{sec:method:filter}:
$\bm{\xi}(t)$ is the augmented state, $\mathbf{H}$ the projection
to the latent value, $(\mathbf{A}_n,\mathbf{Q}_n)$ the discrete
Gauss--Markov transition matrices, and $(\alpha,\beta)$ the
tempering coefficients of the fusion.

\subsection{Spectral Analysis: From the Matérn Kernel to LTI-SDE and Markov Chain}
\label{app:ssgp-spectral}

We take the Matérn kernel as an example to show how to connect a
Gaussian process with a linear time-invariant stochastic
differential equation (LTI-SDE). The Matérn kernel is defined as
\begin{equation}
\kappa_{\nu}(t,t')\;=\;\sigma_{f}^{2}\,\frac{\bigl(\tfrac{\sqrt{2\nu}\,\Delta}{\ell}\bigr)^{\nu}}{\Gamma(\nu)\,2^{\nu-1}}\,K_{\nu}\!\Bigl(\tfrac{\sqrt{2\nu}\,\Delta}{\ell}\Bigr),
\label{eq:matern-kernel-app}
\end{equation}
where $\Delta=|t-t'|$ is the time lag, $\Gamma(\cdot)$ is the
Gamma function, $\sigma_{f}^{2}>0$ is the marginal variance,
$\ell>0$ is the length-scale, $K_{\nu}$ is the modified Bessel
function of the second kind, and $\nu>0$ controls the smoothness
of sample paths from the GP prior $f(t)\!\sim\!\mathcal{GP}\bigl(0,\kappa_{\nu}(t,t')\bigr)$.

For a stationary Matérn kernel
$\kappa_{\nu}(t,t')=\kappa_{\nu}(t-t')$, the energy spectral
density of $f(t)$ can be obtained via the Wiener--Khinchin
theorem by taking the Fourier transform of $\kappa_{\nu}(\Delta)$:
\begin{equation}
S(\omega)\;=\;\frac{q_{s}}{(\lambda^{2}+\omega^{2})^{m+1}},
\label{eq:matern-psd}
\end{equation}
where $\omega$ is the frequency, $q_{s}>0$ is a positive constant
determined by the prefactor of \eqref{eq:matern-kernel-app}, and we
restrict to half-integer smoothness $\nu=m+\tfrac{1}{2}$ for
$m\in\{0,1,2,\ldots\}$. Throughout this appendix we use the
frequency rescaling
\begin{equation}
\lambda\;:=\;\sqrt{2\nu}/\ell,
\label{eq:matern-lambda}
\end{equation}
which keeps the matrices below dimensionally homogeneous. The
denominator of \eqref{eq:matern-psd} is a polynomial of degree
$2(m+1)$ in $\omega$, which is what eventually permits a
finite-dimensional state-space realisation.

Expanding the polynomial $(\lambda+\imath\omega)^{m+1}$ in
\eqref{eq:matern-psd} gives
\begin{equation}
(\lambda+\imath\omega)^{m+1}\;=\;\sum_{k=0}^{m}c_{k}\,(\imath\omega)^{k}\;+\;(\imath\omega)^{m+1},
\label{eq:poly-expansion}
\end{equation}
where the real coefficients $c_{k}$ are given by the binomial
formula $c_{k}=\binom{m+1}{k}\lambda^{m+1-k}$. Since
$(\lambda^{2}+\omega^{2})^{m+1}=\bigl|(\lambda+\imath\omega)^{m+1}\bigr|^{2}$,
\eqref{eq:poly-expansion} allows us to construct an equivalent
frequency-domain system whose output has PSD
\eqref{eq:matern-psd}:
\begin{equation}
\sum_{k=0}^{m}c_{k}\,(\imath\omega)^{k}\,\widehat{f}(\omega)\;+\;(\imath\omega)^{m+1}\,\widehat{f}(\omega)\;=\;\widehat{\beta}(\omega),
\label{eq:freq-system}
\end{equation}
where $\widehat{f}(\omega)$ and $\widehat{\beta}(\omega)$ are the
Fourier transforms of $f(t)$ and of a white-noise process $w(t)$
with spectral density $q_{s}$, respectively.

Taking the inverse Fourier transform of \eqref{eq:freq-system}
returns a stochastic differential equation in the time domain,
\begin{equation}
\sum_{k=0}^{m}c_{k}\,\frac{d^{k}f}{dt^{k}}\;+\;\frac{d^{m+1}f}{dt^{m+1}}\;=\;w(t),
\label{eq:time-sde}
\end{equation}
in which the highest-order derivative of $f$ is driven directly
by the white noise. To remove the high-order derivative, we lift
the scalar process $f(t)$ to a vector-valued state by collecting
$f$ together with its first $m$ time derivatives,
\begin{equation}
\bm{\xi}(t)\;=\;\bigl(f(t),\,f^{(1)}(t),\,\ldots,\,f^{(m)}(t)\bigr)^{\T},
\quad f^{(k)}:=\tfrac{d^{k}f}{dt^{k}},
\label{eq:state-def-app}
\end{equation}
so that $\bm{\xi}(t)\in\R^{m_{\nu}}$ with $m_{\nu}:=m+1$.
Substituting \eqref{eq:state-def-app} into \eqref{eq:time-sde}
turns the high-order SDE into a first-order linear time-invariant
SDE,
\begin{equation}
\frac{d\bm{\xi}(t)}{dt}\;=\;\mathbf{F}\,\bm{\xi}(t)\;+\;\mathbf{L}\,w(t),
\qquad
f(t)\;=\;\mathbf{H}\,\bm{\xi}(t),
\label{eq:lti-sde-canonical}
\end{equation}
in which the drift matrix
$\mathbf{F}\in\R^{m_{\nu}\times m_{\nu}}$ and the noise input
$\mathbf{L}\in\R^{m_{\nu}\times 1}$ take the companion form
\begin{equation}
\mathbf{F}\!=\!
\begin{bmatrix}
0 & 1 &        &   \\
  & \ddots & \ddots &   \\
  &        & 0 & 1 \\
-c_{0} & \cdots & -c_{m-1} & -c_{m}
\end{bmatrix}\!,
\quad
\mathbf{L}\!=\!
\begin{bmatrix}
0\\ \vdots\\ 0\\ 1
\end{bmatrix}\!,
\label{eq:FL-companion}
\end{equation}
together with the constant output projection
$\mathbf{H}=(1,0,\ldots,0)\in\R^{1\times m_{\nu}}$, which extracts
the original process $f(t)$ from the augmented state $\bm{\xi}(t)$.
The companion structure of $\mathbf{F}$ inherits all of its
eigenvalues from the roots of $P(s)=(s+\lambda)^{m+1}$ in
\eqref{eq:matern-psd}; since these roots all lie at $-\lambda<0$,
$\mathbf{F}$ is Hurwitz and \eqref{eq:lti-sde-canonical} is
asymptotically stable.

The LTI-SDE \eqref{eq:lti-sde-canonical} is particularly useful
because its finite-dimensional state $\bm{\xi}(t)$ follows a
Gauss--Markov chain at any set of sampled timestamps. Specifically,
given arbitrary $t_{1}<t_{2}<\cdots<t_{N}$, the joint distribution
of $\{\bm{\xi}(t_{n})\}_{n=1}^{N}$ factorises as
\begin{equation}
p\bigl(\bm{\xi}(t_{1}),\ldots,\bm{\xi}(t_{N})\bigr)\;=\;p(\bm{\xi}(t_{1}))\!\prod_{n=1}^{N-1}\!p(\bm{\xi}(t_{n+1})\mid\bm{\xi}(t_{n})),
\label{eq:joint-factor}
\end{equation}
in which both the initial and the transition distributions are
Gaussian,
\begin{align}
p(\bm{\xi}(t_{1})) &\;=\; \N(\mathbf{0},\,\mathbf{P}_{\infty}),
\label{eq:initial-dist}\\[2pt]
p(\bm{\xi}(t_{n+1})\!\mid\!\bm{\xi}(t_{n})) &\;=\; \N\!\bigl(\mathbf{A}_{n}\bm{\xi}(t_{n}),\,\mathbf{Q}_{n}\bigr).
\label{eq:gauss-markov-chain}
\end{align}
The discrete transition matrices in \eqref{eq:gauss-markov-chain}
are obtained by solving the LTI-SDE
\eqref{eq:lti-sde-canonical} on the inter-frame interval
$\Delta_{n}:=t_{n+1}-t_{n}$, which yields the matrix exponential
\begin{equation}
\mathbf{A}_{n}\;=\;\exp\!\bigl(\Delta_{n}\mathbf{F}\bigr),
\label{eq:An-derivation}
\end{equation}
and the time-integrated process noise
\begin{align}
\mathbf{Q}_{n} &\;=\; \int_{0}^{\Delta_{n}}\!\exp(s\mathbf{F})\,\mathbf{L}\mathbf{L}^{\T}\exp(s\mathbf{F})^{\T}\,q_{s}\,ds
\nonumber\\
&\;=\; \mathbf{P}_{\infty}\;-\;\mathbf{A}_{n}\,\mathbf{P}_{\infty}\,\mathbf{A}_{n}^{\T},
\label{eq:Qn-derivation}
\end{align}
in which the second equality follows from the stationary identity
$\mathbf{P}_{\infty}=\mathbf{A}_{n}\mathbf{P}_{\infty}\mathbf{A}_{n}^{\T}+\mathbf{Q}_{n}$
valid for any stable LTI-SDE. The stationary covariance
$\mathbf{P}_{\infty}$ is the symmetric positive-definite solution
of the algebraic Lyapunov
equation~\cite{lancaster1995algebraic}
\begin{equation}
\mathbf{F}\,\mathbf{P}_{\infty}\;+\;\mathbf{P}_{\infty}\,\mathbf{F}^{\T}\;+\;\mathbf{L}\mathbf{L}^{\T}\,q_{s}\;=\;\mathbf{0},
\label{eq:lyapunov}
\end{equation}
which has a unique solution because $\mathbf{F}$ is Hurwitz, as
noted above. The closed form of $\mathbf{P}_{\infty}$,
$\mathbf{A}_{n}$, and $\mathbf{Q}_{n}$ for the specific case
$\nu=3/2$ used throughout this paper is worked out in
\S\ref{app:matern32-closed-form}.

Equations
\eqref{eq:matern-kernel-app}--\eqref{eq:lyapunov} give a complete
spectral construction of the SS-GP backbone used in
Sec.~\ref{sec:method:filter}: the Matérn covariance is first
Fourier-transformed to the rational PSD \eqref{eq:matern-psd},
which factorises through the polynomial expansion
\eqref{eq:poly-expansion} into a Hurwitz frequency-domain system
\eqref{eq:freq-system}; an inverse Fourier transform brings this
into the time-domain SDE \eqref{eq:time-sde}, which is rewritten
as the first-order LTI-SDE \eqref{eq:lti-sde-canonical} via state
augmentation; and discretising this LTI-SDE on any sequence of
timestamps yields the Gauss--Markov chain
\eqref{eq:joint-factor}--\eqref{eq:Qn-derivation} that the
streaming Kalman recursion of Sec.~\ref{sec:method:filter} operates
on. The construction reduces the cost of joint GP inference from
$\mathcal{O}(N^{3})$ to $\mathcal{O}(N)$~\cite{hartikainen2010kalman}
while preserving the original Matérn covariance exactly at the
sampled times. We close this subsection with a remark on
extensibility: for other stationary kernels (e.g.\ the periodic
kernel), the inverse spectral density $1/S(\omega)$ can be
approximated by a polynomial of $\omega^{2}$ with negative
roots~\cite{solin2014explicit}, after which the same chain of
steps applies.

\subsection{Closed-Form Matrices for Matérn-$3/2$}
\label{app:matern32-closed-form}

With the canonical form of the LTI-SDE \eqref{eq:lti-sde-canonical}
and of the Gauss--Markov chain
\eqref{eq:joint-factor}--\eqref{eq:Qn-derivation} derived above, we
now work out the closed-form matrices for the Matérn-$3/2$ kernel
that this paper uses throughout. The case corresponds to
$\nu=3/2$, equivalently $m=1$ and $m_{\nu}=2$ in
\S\ref{app:ssgp-spectral}, so the augmented state of
\eqref{eq:state-def-app} collapses to the two-dimensional vector
\begin{equation}
\bm{\xi}(t)\;=\;\bigl(f(t),\,\dot{f}(t)\bigr)^{\T}\;\in\;\R^{2},
\label{eq:m32-state}
\end{equation}
i.e.\ each scalar latent dimension $f=\zlat^{(j)}(t)$ of the
Tucker core is tracked together with its time derivative. This
matches the per-dimension definition
$\bm{\xi}^{(j)}(t)=(\zlat^{(j)}(t),\dot{\zlat}^{(j)}(t))^{\T}$
adopted in Sec.~\ref{sec:method:filter}.

Substituting $\nu=3/2$ into the spectral derivation
\eqref{eq:matern-kernel-app}, the Bessel form simplifies to the
familiar Matérn-$3/2$ covariance function
\begin{equation}
\kappa_{3/2}(\tau)\;=\;\sigma_{f}^{2}\Bigl(1+\frac{\sqrt{3}\,\tau}{\ell}\Bigr)\exp\!\Bigl(-\frac{\sqrt{3}\,\tau}{\ell}\Bigr),
\label{eq:m32-kernel}
\end{equation}
where $\tau=|t-t'|$ is the lag. Substituting $m=1$ into the
Hurwitz factor of \eqref{eq:matern-psd} gives
$P(s)=(s+\lambda)^{2}=s^{2}+2\lambda s+\lambda^{2}$ with
$\lambda=\sqrt{3}/\ell$ from \eqref{eq:matern-lambda}, whose
companion form \eqref{eq:FL-companion} specialises to the
$2{\times}2$ drift matrix and $2{\times}1$ noise input
\begin{equation}
\mathbf{F}\;=\;\begin{bmatrix}0 & 1\\ -\lambda^{2} & -2\lambda\end{bmatrix},
\qquad
\mathbf{L}\;=\;\begin{bmatrix}0\\ 1\end{bmatrix},
\label{eq:m32-FL}
\end{equation}
together with the output projection
$\mathbf{H}=(1,0)\in\R^{1\times 2}$ that recovers $f(t)$ from
$\bm{\xi}(t)$. The white-noise spectral density of the equivalent
LTI-SDE is fixed by matching the variance to $\sigma_{f}^{2}$ and
evaluates to
\begin{equation}
q_{s}\;=\;4\,\sigma_{f}^{2}\,\lambda^{3}.
\label{eq:m32-qs}
\end{equation}

For the stationary covariance we substitute
\eqref{eq:m32-FL}--\eqref{eq:m32-qs} into the algebraic Lyapunov
equation \eqref{eq:lyapunov} and solve for the symmetric
$2{\times}2$ matrix $\mathbf{P}_{\infty}$. The off-diagonal
entries cancel by symmetry, and the two diagonal entries decouple
into independent scalar equations, yielding the diagonal closed
form
\begin{equation}
\mathbf{P}_{\infty}\;=\;\begin{bmatrix}\sigma_{f}^{2} & 0\\ 0 & \lambda^{2}\sigma_{f}^{2}\end{bmatrix},
\label{eq:m32-Pinf}
\end{equation}
so that $f$ and $\dot{f}$ are uncorrelated at stationarity with
variances $\sigma_{f}^{2}$ and $3\sigma_{f}^{2}/\ell^{2}$,
respectively. The fact that $\mathbf{P}_{\infty}$ is diagonal is
specific to $\nu=3/2$ and reflects the orthogonality of the
process and its derivative at stationarity for this kernel.

The discrete transition matrix
$\mathbf{A}_{n}=\exp(\Delta_{n}\mathbf{F})$ from
\eqref{eq:An-derivation} admits a known closed form for $\nu=3/2$.
Since the drift matrix $\mathbf{F}$ in \eqref{eq:m32-FL} has a
single eigenvalue $-\lambda$ of multiplicity two, a Jordan
decomposition delivers
\begin{equation}
\mathbf{A}_{n}\;=\;
e^{-\lambda\Delta_{n}}
\begin{bmatrix}
1+\lambda\Delta_{n} & \Delta_{n}\\[2pt]
-\lambda^{2}\Delta_{n} & 1-\lambda\Delta_{n}
\end{bmatrix}.
\label{eq:m32-An}
\end{equation}
Substituting $\mathbf{A}_{n}$ from \eqref{eq:m32-An} and
$\mathbf{P}_{\infty}$ from \eqref{eq:m32-Pinf} into the
process-noise formula \eqref{eq:Qn-derivation} yields
\begin{equation}
\mathbf{Q}_{n}\;=\;\mathbf{P}_{\infty}\;-\;\mathbf{A}_{n}\,\mathbf{P}_{\infty}\,\mathbf{A}_{n}^{\T},
\label{eq:m32-Qn}
\end{equation}
which evaluates to a positive-definite $2{\times}2$ matrix whose
entries depend smoothly on $\Delta_{n}$, $\ell$ and
$\sigma_{f}^{2}$; we do not write out the four entries of
\eqref{eq:m32-Qn} since they appear only inside matrix-level
Kalman operations in Sec.~\ref{sec:method:filter}.

Equations \eqref{eq:m32-kernel}--\eqref{eq:m32-Qn} are the
specific closed-form matrices used by every Kalman and RTS
recursion in this paper. Sec.~\ref{sec:method:filter} places an
independent Matérn-$3/2$ prior on each of the $d$ scalar latent
dimensions $\zlat^{(j)}(t)$ of the Tucker core, so the augmented
state of the joint latent process lives in $\R^{2d}$ and inherits
a block-diagonal
structure: the matrices
$(\mathbf{F},\mathbf{L},\mathbf{H},\mathbf{P}_{\infty},\mathbf{A}_{n},\mathbf{Q}_{n})$
of \eqref{eq:m32-FL}--\eqref{eq:m32-Qn} are replicated along the
diagonal once per latent dimension. This block-diagonal structure
is what makes the per-dimension Kalman recursion of
Sec.~\ref{sec:method:filter} exact rather than an approximation.

\subsection{Tempered PoE Fusion, Augmented-State Update, and the Kalman Corollary}
\label{app:kf-proof}

We now complete the Kalman cycle of
Sec.~\ref{sec:method:filter} by deriving the two operations
introduced informally in the main text: the value-level
tempered fusion of the SS-GP predicted prior with the diffusion
posterior, and the injection of the fused belief back into the
per-dimension augmented state through a Kalman measurement
update. We then show that the composition of these two
operations reduces to the classical Kalman measurement update of
a state-space GP regression at the canonical setting
$\alpha=\beta=1$.

At each streaming frame $n$, TRACE combines two Gaussian beliefs
over the latent value $\zlat_{n}\in\R^{d}$. The first is the
SS-GP predicted prior at the value level,
\begin{equation}
p_{n}^{-}(\zlat_{n})\;=\;\N\bigl(\boldsymbol{\mu}_{n}^{-},\,\boldsymbol{\Sigma}_{n}^{-}\bigr),
\label{eq:value-prior}
\end{equation}
obtained from the per-dimension predicted augmented states
$(\mathbf{m}_{n}^{(j),-},\mathbf{P}_{n}^{(j),-})$ of the SS-GP
forward pass through the output projection
$\mathbf{H}=(1,0)$ derived in
\eqref{eq:lti-sde-canonical}--\eqref{eq:m32-FL}: the $j$-th
component of $\boldsymbol{\mu}_{n}^{-}$ and the
$(j,j)$ entry of $\boldsymbol{\Sigma}_{n}^{-}$ are
\begin{equation}
\mu_{n,j}^{-}\;=\;\mathbf{H}\,\mathbf{m}_{n}^{(j),-},
\qquad
\Sigma_{n,jj}^{-}\;=\;\mathbf{H}\,\mathbf{P}_{n}^{(j),-}\mathbf{H}^{\T},
\label{eq:H-projection}
\end{equation}
and the off-diagonal entries of $\boldsymbol{\Sigma}_{n}^{-}$
vanish because the per-dimension priors are independent. The
second Gaussian belief is the moment-matched diffusion posterior
\begin{equation}
q_{n}^{\mathrm{gen}}(\zlat_{n})\;=\;\N\bigl(\hat{\boldsymbol{\mu}}_{n}^{\mathrm{gen}},\,\hat{\boldsymbol{\Sigma}}_{n}^{\mathrm{gen}}\bigr)
\label{eq:value-diff}
\end{equation}
obtained from the $S$ DPS samples via moment matching.
Substituting the Gaussian forms
\eqref{eq:value-prior} and \eqref{eq:value-diff} into the tempered
PoE rule \eqref{eq:fusion} and computing the
log-density gives the two quadratic forms
\begin{align}
\log p_{n}^{-}(\zlat_{n}) &\;=\; -\tfrac{1}{2}\bigl(\zlat_{n}-\boldsymbol{\mu}_{n}^{-}\bigr)^{\T}(\boldsymbol{\Sigma}_{n}^{-})^{-1}\bigl(\zlat_{n}-\boldsymbol{\mu}_{n}^{-}\bigr)+c_{1},
\label{eq:logprior}\\[2pt]
\log q_{n}^{\mathrm{gen}}(\zlat_{n}) &\;=\; -\tfrac{1}{2}\bigl(\zlat_{n}-\hat{\boldsymbol{\mu}}_{n}^{\mathrm{gen}}\bigr)^{\T}(\hat{\boldsymbol{\Sigma}}_{n}^{\mathrm{gen}})^{-1}\bigl(\zlat_{n}-\hat{\boldsymbol{\mu}}_{n}^{\mathrm{gen}}\bigr)+c_{2},
\label{eq:logdiff}
\end{align}
where $c_{1},c_{2}$ collect constants independent of $\zlat_{n}$.
Taking the tempered linear combination
$\alpha\!\cdot\!\eqref{eq:logprior}+\beta\!\cdot\!\eqref{eq:logdiff}$
and gathering the quadratic and linear terms in $\zlat_{n}$
yields the unnormalised log-density of the fused belief,
\begin{align}
\log q_{n}(\zlat_{n}) \;=\;
&-\tfrac{1}{2}\,\zlat_{n}^{\T}\!\bigl[\alpha(\boldsymbol{\Sigma}_{n}^{-})^{-1}+\beta(\hat{\boldsymbol{\Sigma}}_{n}^{\mathrm{gen}})^{-1}\bigr]\!\zlat_{n} \nonumber\\[2pt]
&+\zlat_{n}^{\T}\!\bigl[\alpha(\boldsymbol{\Sigma}_{n}^{-})^{-1}\boldsymbol{\mu}_{n}^{-}+\beta(\hat{\boldsymbol{\Sigma}}_{n}^{\mathrm{gen}})^{-1}\hat{\boldsymbol{\mu}}_{n}^{\mathrm{gen}}\bigr] \nonumber\\[2pt]
&+\mathrm{const}.
\label{eq:fused-quadratic}
\end{align}
Reading off the precision matrix and natural parameter from
\eqref{eq:fused-quadratic} identifies the fused posterior as a
Gaussian
$q_{n}(\zlat_{n})=\N(\boldsymbol{\mu}_{n},\boldsymbol{\Sigma}_{n})$
with
\begin{align}
\boldsymbol{\Sigma}_{n}^{-1} &\;=\; \alpha\,(\boldsymbol{\Sigma}_{n}^{-})^{-1}\;+\;\beta\,(\hat{\boldsymbol{\Sigma}}_{n}^{\mathrm{gen}})^{-1},
\label{eq:fused-precision}\\[2pt]
\boldsymbol{\Sigma}_{n}^{-1}\boldsymbol{\mu}_{n} &\;=\; \alpha\,(\boldsymbol{\Sigma}_{n}^{-})^{-1}\boldsymbol{\mu}_{n}^{-}\;+\;\beta\,(\hat{\boldsymbol{\Sigma}}_{n}^{\mathrm{gen}})^{-1}\hat{\boldsymbol{\mu}}_{n}^{\mathrm{gen}}.
\label{eq:fused-natural}
\end{align}
Equations \eqref{eq:fused-precision}--\eqref{eq:fused-natural}
are the closed-form fusion invoked in
\eqref{eq:fusion} of the main text. Under the diagonal
fusion default of \S\ref{app:diag-bias}, both
$\boldsymbol{\Sigma}_{n}^{-}$ and
$\hat{\boldsymbol{\Sigma}}_{n}^{\mathrm{gen}}$ are diagonal, and
\eqref{eq:fused-precision}--\eqref{eq:fused-natural} reduce to
$d$ scalar updates of cost $\mathcal{O}(d)$.

The fused belief
$q_{n}(\zlat_{n})=\N(\boldsymbol{\mu}_{n},\boldsymbol{\Sigma}_{n})$
lives at the value level $\zlat_{n}\in\R^{d}$, but the SS-GP
forward chain operates at the augmented-state level
$\bm{\xi}^{(j)}\in\R^{2}$. To close the streaming recursion we
must inject $q_{n}$ back into each per-dimension augmented state
$(\mathbf{m}_{n}^{(j),-},\mathbf{P}_{n}^{(j),-})$. We do so by
treating the $j$-th component of the fused value-level mean
$\boldsymbol{\mu}_{n}$ as a \emph{virtual measurement} of the
scalar latent $\zlat_{n}^{(j)}$ with measurement noise variance
equal to the corresponding diagonal entry of
$\boldsymbol{\Sigma}_{n}$:
\begin{equation}
\mu_{n,j}\;\sim\;\N\!\bigl(\zlat_{n}^{(j)},\,\Sigma_{n,jj}\bigr),
\qquad j=1,\ldots,d.
\label{eq:virtual-meas}
\end{equation}
The $d$ virtual measurements decouple because the diagonal
fusion makes $\boldsymbol{\Sigma}_{n}$ diagonal. The mapping from
the augmented state $\bm{\xi}_{n}^{(j)}\in\R^{2}$ to its scalar
latent value is the same projection $\mathbf{H}=(1,0)$ used in
\eqref{eq:H-projection},
\begin{equation}
\zlat_{n}^{(j)}\;=\;\mathbf{H}\,\bm{\xi}_{n}^{(j)},
\label{eq:state-projection}
\end{equation}
so \eqref{eq:virtual-meas} together with \eqref{eq:state-projection}
defines a linear Gaussian measurement model on each augmented
state with measurement matrix $\mathbf{H}$ and noise variance
$\Sigma_{n,jj}$.

Applying the standard Kalman measurement update of
$(\mathbf{m}_{n}^{(j),-},\mathbf{P}_{n}^{(j),-})$ under this
linear Gaussian observation produces the per-dimension
augmented-state update of Sec.~\ref{sec:method:filter}. The Kalman
gain is
\begin{equation}
\mathbf{K}_{n}^{(j)}\;=\;\mathbf{P}_{n}^{(j),-}\,\mathbf{H}^{\T}\!\bigl(\mathbf{H}\,\mathbf{P}_{n}^{(j),-}\mathbf{H}^{\T}+\Sigma_{n,jj}\bigr)^{\!-1},
\label{eq:Kalman-gain}
\end{equation}
the updated state mean is
\begin{equation}
\mathbf{m}_{n}^{(j)}\;=\;\mathbf{m}_{n}^{(j),-}\;+\;\mathbf{K}_{n}^{(j)}\bigl(\mu_{n,j}-\mathbf{H}\,\mathbf{m}_{n}^{(j),-}\bigr),
\label{eq:Kalman-mean}
\end{equation}
and the updated state covariance is
\begin{equation}
\mathbf{P}_{n}^{(j)}\;=\;\bigl(\mathbf{I}-\mathbf{K}_{n}^{(j)}\mathbf{H}\bigr)\mathbf{P}_{n}^{(j),-}.
\label{eq:Kalman-cov}
\end{equation}
Equations \eqref{eq:Kalman-gain}--\eqref{eq:Kalman-cov} give the
per-dimension form of the augmented-state lift
described in Sec.~\ref{sec:method:filter}. The cost of one update is
constant in $d$ because all matrices in
\eqref{eq:Kalman-gain}--\eqref{eq:Kalman-cov} are
$2{\times}2$, and the $d$ updates are run independently for the
$d$ latent dimensions, so the total per-frame cost of the
augmented-state update is $\mathcal{O}(d)$. We can simplify
\eqref{eq:Kalman-gain} further by substituting
$\mathbf{H}=(1,0)$: the innovation variance
$\mathbf{H}\mathbf{P}_{n}^{(j),-}\mathbf{H}^{\T}+\Sigma_{n,jj}$
is the scalar $\bigl(\mathbf{P}_{n}^{(j),-}\bigr)_{\!1,1}+\Sigma_{n,jj}$,
and the Kalman gain $\mathbf{K}_{n}^{(j)}\in\R^{2\times 1}$ is
the first column of $\mathbf{P}_{n}^{(j),-}$ divided by this
scalar.

At the canonical setting $\alpha=\beta=1$, the composition of
the value-level fusion
\eqref{eq:fused-precision}--\eqref{eq:fused-natural} with the
augmented-state update
\eqref{eq:Kalman-gain}--\eqref{eq:Kalman-cov} reduces exactly to
the classical Kalman measurement update of a state-space GP
regression. To see this, set $\alpha=\beta=1$ in
\eqref{eq:fused-precision}--\eqref{eq:fused-natural}:
\begin{align}
\boldsymbol{\Sigma}_{n}^{-1} &\;=\; (\boldsymbol{\Sigma}_{n}^{-})^{-1}\;+\;(\hat{\boldsymbol{\Sigma}}_{n}^{\mathrm{gen}})^{-1},
\label{eq:KF-precision}\\[2pt]
\boldsymbol{\Sigma}_{n}^{-1}\boldsymbol{\mu}_{n} &\;=\; (\boldsymbol{\Sigma}_{n}^{-})^{-1}\boldsymbol{\mu}_{n}^{-}\;+\;(\hat{\boldsymbol{\Sigma}}_{n}^{\mathrm{gen}})^{-1}\hat{\boldsymbol{\mu}}_{n}^{\mathrm{gen}}.
\label{eq:KF-natural}
\end{align}
Equations \eqref{eq:KF-precision}--\eqref{eq:KF-natural} are
exactly the information-form Kalman measurement update that
treats $q_{n}^{\mathrm{gen}}$ as a Gaussian observation
$\mathbf{y}=\hat{\boldsymbol{\mu}}_{n}^{\mathrm{gen}}$ on the
value-level prior
$p_{n}^{-}(\zlat_{n})=\N(\boldsymbol{\mu}_{n}^{-},\boldsymbol{\Sigma}_{n}^{-})$
with measurement matrix $\mathbf{I}_{d}$ and noise covariance
$\hat{\boldsymbol{\Sigma}}_{n}^{\mathrm{gen}}$. Applying the
Woodbury identity to \eqref{eq:KF-precision} returns the
equivalent covariance-form update
\begin{equation}
\boldsymbol{\Sigma}_{n}\;=\;\boldsymbol{\Sigma}_{n}^{-}\;-\;\boldsymbol{\Sigma}_{n}^{-}\!\bigl(\boldsymbol{\Sigma}_{n}^{-}+\hat{\boldsymbol{\Sigma}}_{n}^{\mathrm{gen}}\bigr)^{\!-1}\!\boldsymbol{\Sigma}_{n}^{-}.
\label{eq:woodbury-update}
\end{equation}
Substituting \eqref{eq:KF-precision}--\eqref{eq:woodbury-update}
into the per-dimension augmented-state update
\eqref{eq:Kalman-gain}--\eqref{eq:Kalman-cov} reproduces the
textbook Kalman recursion of an SS-GP regression in which the
``observation'' at frame $n$ is the moment-matched DPS Gaussian
summary $q_{n}^{\mathrm{gen}}$. For readers familiar with state-space GP
regression, this equivalence is visible directly from
\eqref{eq:KF-precision}; we record it explicitly only to fix the
correspondence between TRACE's tempered fusion plus
augmented-state update and the classical Kalman terminology.

For $(\alpha,\beta)\neq(1,1)$, the composition
\eqref{eq:fused-precision}--\eqref{eq:Kalman-cov} is a strict
generalisation of the Kalman measurement update. The exponents
act as trust weights on each Gaussian source: smaller $\alpha$
down-weights the SS-GP prior under fast dynamics where the
$\kappa(\Delta)$ contraction of \S\ref{app:matern32-closed-form}
is too tight; smaller $\beta$ down-weights the diffusion
posterior when $q_{n}^{\mathrm{gen}}$ is under-determined by a near-empty
observation. The validity of the closed form is unchanged:
equations \eqref{eq:fused-precision}--\eqref{eq:Kalman-cov} stay
well-defined and positive-definite for any
$\alpha,\beta>0$, so the same per-dimension augmented-state
update is reused under any tempering schedule.

\subsection{Diagonal Fusion: Rationale and Cost}
\label{app:diag-bias}

The diagonal fusion of Sec.~\ref{sec:method:filter} replaces the
moment-matched diffusion covariance
$\hat{\boldsymbol{\Sigma}}_{n}^{\mathrm{gen}}$ entering
\eqref{eq:fusion} by its diagonal
$\mathbf{D}:=\mathrm{diag}(\hat{\boldsymbol{\Sigma}}_{n}^{\mathrm{gen}})$.
The motivation is both statistical and computational. Because the number
of DPS samples is far smaller than the latent dimension ($S\ll d$), the
empirical full covariance estimated from the $S$ samples is
rank-deficient (rank at most $S-1$) and numerically unstable to invert in
the high-dimensional latent space. The diagonal moment approximation
avoids this inversion, reducing the per-frame fusion cost from
$\mathcal{O}(d^{3})$ to $\mathcal{O}(d)$ and remaining stable at the
Supernova scale ($d=32{,}768$), where forming and inverting a dense
$d\times d$ precision matrix is infeasible. We examine its empirical
effect---including the instability of the full-covariance variant---in
Appendix~\ref{app:extra} (Table~\ref{tab:fusion}).

\subsection{RTS Smoother and Off-Grid Bridge Interpolation}
\label{app:rts-bridge}\label{app:rts}\label{app:bridge}

After the forward filter terminates with the augmented-state means
and covariances $\{(\mathbf{m}_{n}^{(j)},\mathbf{P}_{n}^{(j)})\}_{n=1}^{N}$
for each latent dimension $j$, the Rauch--Tung--Striebel
smoother~\cite{rauch1965maximum,sarkka2019applied} delivers the
smoothed posterior in a single backward sweep of cost
$\mathcal{O}(N)$ per dimension. Initialising at $n=N$ with the
terminal filter mean and covariance and recursing backward, the
update is
\begin{align}
\mathbf{G}_{n}^{(j)} &\;=\; \mathbf{P}_{n}^{(j)}\,\mathbf{A}_{n+1}^{\T}\,\bigl(\mathbf{P}_{n+1}^{(j),-}\bigr)^{-1},
\label{eq:rts-gain}\\[2pt]
\mathbf{m}_{n}^{(j),s} &\;=\; \mathbf{m}_{n}^{(j)} \;+\; \mathbf{G}_{n}^{(j)}\bigl(\mathbf{m}_{n+1}^{(j),s}-\mathbf{m}_{n+1}^{(j),-}\bigr),
\label{eq:rts-mean}\\[2pt]
\mathbf{P}_{n}^{(j),s} &\;=\; \mathbf{P}_{n}^{(j)} \;+\; \mathbf{G}_{n}^{(j)}\bigl(\mathbf{P}_{n+1}^{(j),s}-\mathbf{P}_{n+1}^{(j),-}\bigr)\mathbf{G}_{n}^{(j),\T}.
\label{eq:rts-cov}
\end{align}
The smoothed augmented states are then projected through
$\mathbf{H}$ to recover the smoothed latent-value posterior at
every grid time $t_{n}$, supplying the RTS-corrected estimates
used by Sec.~\ref{sec:method:filter}.

For an unseen query time
$t^{\star}\!\in\!(t_{k},t_{k+1})$, the latent value
$\zlat(t^{\star})$ is not directly produced by the filter or the
smoother; we must instead infer it from the smoothed boundary
states $(\boldsymbol{\mu}_{k},\boldsymbol{\mu}_{k+1})$ at the
neighbouring grid times. The Markov property of the SS-GP gives
this inference a closed form, derived below.

Markovianity factorises the joint density of
$(\bm{\xi}_{k},\bm{\xi}^{\star},\bm{\xi}_{k+1})$ given
$\Obshist$ into two Gauss--Markov transitions of the form
\eqref{eq:gauss-markov-chain}:
\begin{align}
p(\bm{\xi}^{\star}\!\mid\!\bm{\xi}_{k}) &\;=\; \N\!\bigl(\mathbf{A}_{1}\bm{\xi}_{k},\,\mathbf{Q}_{1}\bigr),
\label{eq:bridge-first}\\[2pt]
p(\bm{\xi}_{k+1}\!\mid\!\bm{\xi}^{\star}) &\;=\; \N\!\bigl(\mathbf{A}_{2}\bm{\xi}^{\star},\,\mathbf{Q}_{2}\bigr),
\label{eq:bridge-second}
\end{align}
where the matrices $(\mathbf{A}_{i},\mathbf{Q}_{i})$ are obtained
from \eqref{eq:m32-An}--\eqref{eq:m32-Qn} at the sub-intervals
$\Delta_{1}=t^{\star}-t_{k}$ and
$\Delta_{2}=t_{k+1}-t^{\star}$, respectively.

Conditioning the joint
\eqref{eq:bridge-first}--\eqref{eq:bridge-second} on the smoothed
boundary states
$(\boldsymbol{\mu}_{k},\boldsymbol{\mu}_{k+1})$ and reading off the
quadratic in $\bm{\xi}^{\star}$ produces a Gaussian
$q(\zlat(t^{\star}))=\N(\boldsymbol{\mu}^{\star},\mathbf{V}^{\star})$.
Completing the square gives the precision and natural parameter
in closed form,
\begin{align}
\bigl(\mathbf{V}^{\star}\bigr)^{-1} &\;=\; \mathbf{Q}_{1}^{-1} \;+\; \mathbf{A}_{2}^{\T}\,\mathbf{Q}_{2}^{-1}\,\mathbf{A}_{2},
\label{eq:bridge-precision}\\[4pt]
\bigl(\mathbf{V}^{\star}\bigr)^{-1}\boldsymbol{\mu}^{\star} &\;=\; \mathbf{Q}_{1}^{-1}\,\mathbf{A}_{1}\,\boldsymbol{\mu}_{k} \;+\; \mathbf{A}_{2}^{\T}\,\mathbf{Q}_{2}^{-1}\,\boldsymbol{\mu}_{k+1},
\label{eq:bridge-natural}
\end{align}
from which the field-level prediction of Eq.~\eqref{eq:field-query} is obtained.

Composing the bridge mean $\boldsymbol{\mu}^{\star}$ from
\eqref{eq:bridge-natural} with the FTM decoder $\Gobs(\rstar)$
delivers a field prediction at the arbitrary off-grid query
$(\rstar,t^{\star})$:
\begin{equation}
\hat{\fieldy}(\rstar,t^{\star}) \;=\; \Gobs(\rstar)^{\T}\,\boldsymbol{\mu}^{\star},
\label{eq:bridge-field}
\end{equation}
in agreement with the field query of Eq.~\eqref{eq:field-query} in \S\ref{sec:method:smooth}.


\section{Dataset Sources and Physical Context}
\label{app:datasets}

\begin{table}[h]
\centering\small
\caption{Per-dataset architecture configuration: Tucker
multilinear rank $R$, vectorised-core latent dimension
$d=\prod_k R_k$, and SIREN basis frequency $\omega$.}
\label{tab:arch_config}
\begin{tabular}{lccc}
\toprule
 & \textbf{AM} & \textbf{Ocean} & \textbf{SN} \\
\midrule
Tucker rank $R$  & $(1,48,48)$ & $(5,20,20)$ & $(32,32,32)$ \\
Latent dim. $d$  & $2304$ & $2000$ & $32768$ \\
SIREN $\omega$   & $20$ & $10$ & $20$ \\
\bottomrule
\end{tabular}
\end{table}

This section provides physical context, data sources, and
dataset-level statistics for the three benchmarks.

\paragraph{Active Matter (AM).}
A two-dimensional active-nematic continuum field describing the
collective dynamics of rod-like active particles suspended in a Stokes
fluid. Each frame is a scalar field over the unit square
($256{\times}256$, $T{=}24$).  We use $900$ training trajectories and
$28$ held-out test trajectories, with a fixed $10\%$ training mask.
Data are obtained from the \emph{active\_matter} collection of The
Well.\footnote{\url{https://polymathic-ai.org/the_well/datasets/active_matter/}}

\paragraph{Ocean.}
Pacific sound-speed reanalysis fields covering five depth layers,
derived from the HYCOM operational reanalysis
product.\footnote{\url{https://www.hycom.org/}} Each frame is a
layered scalar field over a fixed latitude--longitude grid
($5{\times}38{\times}76$, $T{=}24$).  We use $950$ training
trajectories and $50$ held-out test trajectories, with a fixed
$10\%$ training mask.

\paragraph{Supernova (SN).}
Three-dimensional temperature evolution of a supernova blast wave
propagating through a dense monatomic ideal-gas cloud
($64^3$, $T{=}16$).  We use $370$ training trajectories and $26$
held-out test trajectories, with a fixed $15\%$ training mask.  Data
are obtained from the \emph{supernova\_explosion\_64} collection of
The Well.\footnote{\url{https://polymathic-ai.org/the_well/datasets/supernova_explosion_64/}}


\section{Implementation and Hyperparameters}
\label{app:impl}

\begin{table}[t]
\centering\small
\caption{Inference hyperparameters per dataset and observation regime.
Shared across all runs: $S{=}20$, $N{=}100$,
$\sigma_{\text{obs}}{=}0.05$, $\beta{=}1$, diagonal fusion.}
\label{tab:hparams}

\begin{tabular}{llcc}
\toprule
Dataset & Regime & $\zeta$ & $(\alpha,\ell,\sigma_f)$ \\
\midrule
\multirow{3}{*}{AM}
 & Control             & $0.001$ & $(0.5,1.0,2.0)$ \\
 & Temporally Sparse   & $0.001$ & $(0.5,5.0,1.0)$ \\
 & Spatially Localized & $0.01$  & $(0.5,5.0,1.0)$ \\
\midrule
\multirow{3}{*}{Ocean}
 & Control             & $0.06$  & $(0.5,2.0,1.0)$ \\
 & Temporally Sparse   & $0.06$  & $(0.5,5.0,2.0)$ \\
 & Spatially Localized & $0.2$   & $(0.5,5.0,2.0)$ \\
\midrule
\multirow{3}{*}{SN}
 & Control             & $0.02$  & $(0.5,2.0,1.0)$ \\
 & Temporally Sparse   & $0.02$  & $(0.3,5.0,1.0)$ \\
 & Spatially Localized & $0.02$  & $(0.7,20,2.0)$ \\
\bottomrule
\end{tabular}
\end{table}

All models are trained and evaluated with PyTorch~2.9 on a single
NVIDIA RTX~5090 (32\,GB). EDM training uses FP32 throughout; FP16
diverges in the score network at our scale.

\subsection{FTM Pretraining}
\label{app:impl:ftm}

The shared continuous-coordinate basis
$\{\boldsymbol{\phi}^{(k)}_{\theta_k}\}$ is a per-mode
SIREN~\cite{sitzmann2020siren} with frequency $\omega$, jointly
trained with the per-trajectory Tucker cores by minimising the
masked reconstruction loss on the training split (mask ratio per
dataset reported in Appendix~\ref{app:datasets}).  The full training
objective is
\begin{multline}
\mathcal{L}_{\mathrm{FTM}}
=
\mathbb{E}_{(\rrr,t,y)\sim\Datacorpus}\!
\bigl\|y-\hat{\fieldy}(\rrr,t)\bigr\|_{2}^{2}
\\
+\;\beta_{\mathrm{TV}}\,
\mathbb{E}_{b}\!\sum_{n=2}^{N^{(b)}}
\|\Wcore_n-\Wcore_{n-1}\|_{F}^{2},
\label{eq:ftm-loss-app}
\end{multline}
where the first term is evaluated only at observed off-grid
coordinates, and the temporal TV regulariser penalises frame-to-frame
drift of the Tucker core in the Frobenius norm with weight
$\beta_{\mathrm{TV}}=10^{-7}$.  The basis parameters and cores are
alternately updated by
Adam with learning rate
$2\times10^{-4}$.  Per-dataset multilinear ranks $R$ and SIREN
frequencies $\omega$ are reported in
Table~\ref{tab:arch_config}.  The basis is frozen after
pretraining and reused by all functional-Tucker methods.

\subsection{Latent Diffusion (EDM) Pretraining}
\label{app:impl:edm}

We adopt the EDM framework~\cite{karras2022elucidating} with an identity  
reducer (the latent is the vectorised Tucker core).  The denoiser
$D_{\boldsymbol{\theta}}$ is a Song-UNet~\cite{song2021score} internally
parameterised over the tensor form
$\Wcore_n=\operatorname{vec}^{-1}(\zlat_n)$ to preserve the mode-wise
spatial structure of the Tucker core; the training objective is the
standard EDM weighted reconstruction loss
\begin{equation}
\mathcal{L}_{\mathrm{EDM}}
=
\mathbb{E}_{\zlat,\sigma,\boldsymbol{\varepsilon}}\!
\Bigl[\lambda(\sigma)\,
\bigl\|D_{\boldsymbol{\theta}}(\zlat+\sigma\boldsymbol{\varepsilon};
\sigma)-\zlat\bigr\|_{2}^{2}\Bigr],
\label{eq:edm-loss-app}
\end{equation}
with noise-level weighting $\lambda(\sigma)=(\sigma^{2}+\sigma_{\mathrm{data}}^{2})/(\sigma\cdot\sigma_{\mathrm{data}})^{2}$~\cite{karras2022elucidating}.  The cores are $z$-scored before
training; critically, $\sigma_{\text{data}}$ is read from the
post-normalisation training statistics
($\sigma_{\text{data}}\approx1.0$) rather than from the raw core
standard deviation, as the latter produces large reconstruction errors.
Training uses Adam (learning rate $2\times10^{-4}$, batch size $128$,
EMA half-life $500$\,kimg) for $500$ epochs.  All training is in FP32;
FP16 diverges in the score network at our scale.

\subsection{SS-GP and DPS Inference Setup}
\label{app:impl:ssgp}

The temporal prior is a state-space Gaussian process with a Mat\'ern-$3/2$
kernel, parameterised by a length-scale $\ell$ and marginal standard
deviation $\sigma_f$, combined with the diffusion prior through a
tempered closed-form Gaussian fusion with coefficients $(\alpha,\beta)$.
Throughout we fix the diffusion-side temper $\beta=1$ and tune only the
prior-side temper $\alpha$, so $\alpha$ alone controls how strongly the
temporal prior is trusted ($\alpha=0$ recovers per-frame DPS). We use the
diagonal fusion variant throughout (\S\ref{app:exp:ablation} compares
fusion variants). The length-scale $\ell$ governs how strongly past
frames inform the present: a short $\ell$ suffices in the control regime
where every frame is observed, whereas a longer $\ell$ is essential to
bridge the unobserved intervals of the temporally sparse and spatially
localized regimes. This dependence is not oracle tuning: $\ell$ is set
from the \emph{observable} missing rate rather than the target, and
Fig.~\ref{fig:sens} shows performance is governed by $\ell$ with a broad
plateau (so a single $\ell{=}5$ is near-optimal across all sparse
settings) and is largely insensitive to $\alpha$.

For the per-frame measurement injection we use the batched diffusion
posterior sampling~\cite{chung2023dps} implementation, which batches  
all $S=20$ posterior samples in one network call at each of the $N=100$
reverse-diffusion steps, measurement-noise scale $\sigma_{\text{obs}}=0.05$, and
a guidance step size $\zeta$ set per dataset and observation regime
(Table~\ref{tab:hparams}). The sample count $S=20$ is chosen to sit on
the accuracy plateau of the Gaussian moment summary. In a controlled
ablation on Active Matter (Control regime, $\rho=1\%$, a $10$-trajectory
development subset, identical configuration), the field-domain RMSE is
$0.083$ at $S=5$, $0.081$ at $S=20$, and $0.080$ at $S=50$ (the $S{=}50$
point combines the cached $20$ samples with $30$ freshly drawn independent
ones): increasing $S$ beyond $20$ changes RMSE by under $0.0005$, i.e.\
well inside the $\approx\!0.03$ cross-trajectory standard deviation. (The
$0.081$ here is on this $10$-trajectory subset; the $0.078$ reported for
TRACE-Frame in the control-regime table (Table~\ref{tab:control}) is over
the full $28$-trajectory test set at the same $S=20$.) We therefore keep
$S=20$, which retains the second-moment information while keeping cost
linear in $S$; the same negligible $S{=}20\!\to\!50$ gain holds on
Supernova and the moving-window setting.

\subsection{Per-Dataset Hyperparameter Table}
\label{app:impl:hparams}

Table~\ref{tab:hparams} summarises the inference hyperparameters. The
DPS guidance step $\zeta$ is fixed once per dataset and regime.
The tempering coefficient $\alpha$ is set to $0.5$ as the default,
which engages the SS-GP transition prior; the two Supernova settings
deviate from this default ($\alpha=0.3$ under the temporal regime and
$\alpha=0.7$ for the spatial moving-cube) due to the weak
frame-to-frame dynamics of the Supernova field. The length-scale
$\ell$ is the principal parameter we increase for missing/blackout and
moving-window settings, so that the prior can bridge longer unobserved
intervals. Setting $\alpha=0$ would disable the transition prior and
reduce the method to per-frame DPS. In our experiments hyperparameters
are fixed per evaluation setting to allow controlled comparison; in
deployment one would either choose the conservative regime (longer
$\ell$) or estimate the missing-frequency regime online --- we leave
this adaptive tuning to future work.

\subsection{Moving-Window Trajectories}
\label{app:impl:window}

Figure~\ref{fig:window_traj} illustrates the three trajectory types on
the Active Matter domain. Each dataset uses the same three patterns
(S-curve, circular 1-loop, circular 3-loop) adapted to its
dimensionality, with local observation density
$\rho_{\text{loc}}=15\%$ inside the window on all datasets. Full sweep
ranges (up to 6 loops) are documented alongside results in
Appendix~\ref{app:extra}.

\noindent\textbf{Active Matter} ($256{\times}256$, $T{=}24$; window
$96{\times}96$). The S-curve follows a serpentine raster: the window
sweeps left-to-right along one axis, steps down by half a window
width, and reverses direction, covering the domain in a single pass.
The circular trajectory moves the window centre along a closed
elliptical orbit; each additional lap revisits previously-covered
regions. The main text reports 1- and 3-lap results.

\noindent\textbf{Ocean} ($5{\times}38{\times}76$, $T{=}24$; 3D window
$5{\times}24{\times}24$, i.e.\ the full 5-layer depth and a
$24{\times}24$ lateral tile sliding on the $38{\times}76$ lateral
plane). The window centre follows the same serpentine and elliptical
patterns in the lateral dimensions while all five depth layers are
observed at every position.

\subsubsection{Supernova 3D Moving-Window Trajectories}
For Supernova, the moving window is a $32^3$ cube within the $64^3$
domain. The cube centre follows a three-dimensional trajectory with
two variants. In the 3D S-curve, the cube scans the W--D plane in a
serpentine pattern while advancing linearly through the H dimension;
the number of rows is determined by the cube size and domain width.
In the 3D circular trajectory, the cube centre traces an elliptical
orbit in the W--D plane while the H coordinate follows a sinusoidal
bounce at half the orbital frequency, producing a helical path.
The number of loops $n_{\text{loops}}$ controls the angular velocity
($2\pi n_{\text{loops}}/T$) and thus the revisit frequency; the main
text reports $n_{\text{loops}}=1,3$, with sweeps up to $6$ in the
appendix. At $T=16$ frames, one loop yields approximately $6$ pixels
of lateral displacement per frame.


\section{Algorithmic Pseudocode}
\label{app:algorithm}

Algorithms~\ref{alg:trace_pretrain} and
\ref{alg:trace_infer} outline the offline pretraining and online
streaming inference procedures of TRACE. All symbols follow the
notation of Section~\ref{sec:method}; implementation details are
provided in Appendix~\ref{app:impl}.

\begin{algorithm}[tb]
\caption{Offline Pretraining of TRACE}
\label{alg:trace_pretrain}
\textbf{Input:} Training corpus $\Datacorpus=\{\Obs^{(b)}\}_{b=1}^{B}$;
  multilinear ranks $\{R_k\}_{k=1}^{K}$; EDM noise schedule $p(\sigma)$.\\[2pt]
\textbf{Output:} Frozen FTM basis $\{\boldsymbol{\phi}^{(k)}_{\theta_k}\}_{k=1}^{K}$
  and pretrained latent denoiser $D_{\boldsymbol{\theta}}$.
\begin{algorithmic}[1]
\STATE \textbf{Stage I: Continuous latent representation learning.}
\STATE \quad Initialise basis $\{\theta_k\}$ and cores $\{\Wcore_n^{(b)}\}_{b,n}$.
\REPEAT
  \STATE Sample minibatch $(\rrr,t,y)\sim\Datacorpus$;
          reconstruct $\hat{\fieldy}(\rrr,t)$ via Eq.~\eqref{eq:ftm-entry};
          update by minimising $\mathcal{L}_{\mathrm{FTM}}$.
\UNTIL{convergence}
\STATE Freeze basis; construct latent dataset
  $\mathcal{Z}=\{\zlat_n^{(b)}=\mathrm{vec}(\Wcore_n^{(b)})\}_{b,n}$.
\STATE \textbf{Stage II: Latent diffusion prior learning.}
\REPEAT
  \STATE Sample $\zlat\sim\mathcal{Z}$, $\sigma\sim p(\sigma)$,
          $\boldsymbol{\varepsilon}\sim\mathcal{N}(\mathbf{0},\mathbf{I})$;
          update $D_{\boldsymbol{\theta}}$ by minimising
          $\mathcal{L}_{\mathrm{EDM}}$.
\UNTIL{convergence}
\RETURN $\{\boldsymbol{\phi}^{(k)}_{\theta_k}\}$ and $D_{\boldsymbol{\theta}}$.
\end{algorithmic}
\end{algorithm}

\begin{algorithm}[tb]
\caption{TRACE Streaming Inference}
\label{alg:trace_infer}
\textbf{Input:} Pretrained FTM basis and EDM denoiser;
  streaming observations $\{\Obs_n\}_{n=1}^{N}$;
  SS-GP parameters $(\sigma_f,\ell)$; tempering $(\alpha,\beta)$;
  DPS settings $(S,N_\sigma,\zeta)$.\\[2pt]
\textbf{Output:} Filtering posteriors $\{q_n^{\mathrm{filt}}\}_{n=1}^{N}$,
  smoothed posteriors $\{q_n^{\mathrm{smooth}}\}_{n=1}^{N}$,
  continuous predictor $\hat{\fieldy}(\rstar,t^\star)$.
\begin{algorithmic}[1]
\STATE \textbf{Initialisation.} Set augmented state $(\mathbf{m}_0,\mathbf{P}_0)$
  at stationary prior.

\STATE \textbf{Forward filtering pass.}
\FOR{$n=1,\ldots,N$}
  \STATE SS-GP predict: propagate $(\mathbf{m}_{n-1},\mathbf{P}_{n-1})$
          $\rightarrow$ $(\mathbf{m}_n^{-},\mathbf{P}_n^{-})$
          $\rightarrow$ $p_n^{-}(\zlat_n)$ (Eq.~\eqref{eq:pred-prior}).
  \IF{$\Obs_n\neq\emptyset$}
    \STATE DPS sampling (Eq.~\eqref{eq:dps-trace}) $\rightarrow$
            $S$ samples $\{\zlat_n^{(s)}\}$;
            moment-match $\rightarrow$ $q_n^{\mathrm{gen}}$
            (Eq.~\ref{eq:moment-match});
            tempered fusion (Eq.~\eqref{eq:fusion})
            $\rightarrow$ $q_n^{\mathrm{filt}}$.
  \ELSE
    \STATE $q_n^{\mathrm{filt}}\leftarrow p_n^{-}$ (missing frame).
  \ENDIF
  \STATE Moment-consistent lift to augmented state
          (Appendix~\ref{app:kf-proof})
          $\rightarrow$ $(\mathbf{m}_n,\mathbf{P}_n)$.
\ENDFOR

\STATE \textbf{Backward smoothing pass.}
\FOR{$n=N-1,\ldots,1$}
  \STATE RTS update: $\mathbf{m}_n^{\mathrm{s}},\mathbf{P}_n^{\mathrm{s}}$
          (see Appendix~\ref{app:rts}).
\ENDFOR
\STATE Project to smoothed latent posteriors $\{q_n^{\mathrm{smooth}}\}$.

\STATE \textbf{Continuous prediction.}
  For any $(\rstar,t^\star)$, SS-GP bridge
  (Appendix~\ref{app:bridge}) $\rightarrow$
  $\hat{\fieldy}(\rstar,t^\star)=\Gobs(\rstar)^{\T}\boldsymbol{\mu}^\star$.
\RETURN $\{q_n^{\mathrm{filt}}\}$, $\{q_n^{\mathrm{smooth}}\}$, continuous predictor.
\end{algorithmic}
\end{algorithm}


\section{Baseline Reproduction Details}
\label{app:baselines}

We detail how each baseline is reproduced, the adaptations required to
apply it to off-grid sparse field reconstruction, and the per-method
tuning protocol. All methods observe identical scattered masks under
each setting and are evaluated on the same held-out trajectories, with
each baseline tuned over its principal hyperparameters.

\subsection{Shared-Basis Protocol}
The tensor- and diffusion-based reconstructors are placed on the same
footing as our framework by decoding through the \emph{same} pretrained
continuous-coordinate basis $\{\boldsymbol{\phi}^{(k)}_{\theta_k}\}$.
Concretely, \textbf{LRTFR} solves only for the per-frame Tucker cores
under a temporal total-variation penalty while keeping our frozen basis
fixed (the basis frequency $\omega$ matches our FTM pretraining), and
\textbf{SDIFT} decodes its sampled cores through the identical basis.
This isolates the inference algorithm--temporal total variation
(LRTFR), batch GP-sequential diffusion (SDIFT), or our streaming
SS-GP prior--as the only variable. SDIFT additionally trains its own
latent diffusion network on its own core statistics; the resulting core
distribution differs from ours (nonzero mean after normalisation),
which we report as a reproduction property rather than a tuning artifact.
\textbf{MMGN} and \textbf{DBF} do not use a Tucker
factorisation and retain their native representations.

\subsection{MMGN and the Interpolation Extension}
MMGN fits an implicit neural field independently at each frame and, by
construction, produces a meaningful reconstruction only for frames that
carry observations; on a frame without measurements its per-frame latent
is undetermined. To extend MMGN to unobserved frames---rather than
penalise it with an empty output---we exploit the temporal continuity of
its latent code and linearly interpolate the codes of the nearest
observed frames. This extension \emph{strengthens} the baseline: on
Active Matter ($\rho = 1\%$), interpolation reduces the RMSE under
missing-frame settings from $0.562$ to $0.187$ and from $0.762$ to
$0.313$ as the missing interval widens, relative to the native
zero-latent variant. We therefore report the interpolated variant as the
stronger MMGN baseline throughout.

\subsection{SDIFT}
SDIFT is trained following its released configuration (effective batch
size $16$ via gradient accumulation) and its message-passing posterior
sampling (MPDPS) is tuned over its guidance strength, step count, and
$\zeta$ schedule. For the 3D Supernova field we adapt its
sampling pipeline to the cubic moving-window observations. As noted
above, SDIFT uses its own pretrained diffusion network; the distribution
shift relative to our cores partly explains its weaker field-domain
accuracy and is not a scaling artifact.

\subsection{DBF and the Observation-Density Protocol}
The deep Bayesian filter (DBF) is an end-to-end causal filter that does
not use a Tucker factorisation. Two adaptations were required for a
fair reproduction. First, its decoder applies a terminal
$\mathrm{ReLU}$, which the original DBF paper validates on strictly
non-negative fluid simulations and therefore reflects a
domain-specific representation choice rather than a
method-essential constraint; on our zero-centred fields the same
$\mathrm{ReLU}$ collapses the output to the mean field, so we remove
this activation and $z$-score the targets, after which DBF produces
non-trivial reconstructions (its observation-driven correlation
substantially exceeds that of a mean-field template).
Second, DBF requires a minimum training density. Trained at our
evaluation density $\rho = 1\%$, optimisation \emph{does} proceed but
converges to a degenerate solution: a near-constant mean-field predictor
(RMSE $\approx 1.1$, no better than predicting the dataset mean), so no
useful filter is learned. We therefore train it at the lowest density at
which it converges to a genuine, observation-dependent reconstruction
($\rho = 3\%$) and evaluate across all test densities. This necessarily
introduces a train/test density mismatch (train $\rho=3\%$, test
$\rho=1\%$). To show the mismatch does not unfairly penalise DBF,
Table~\ref{tab:dbf_matched} additionally reports DBF trained \emph{and}
tested at $\rho = 3\%$---its strongest attainable setting. Even there it
trails our framework by a wide margin with full temporal availability
(e.g.\ $0.473$ vs.\ $0.066$ on Active Matter, a $7.1\times$ gap, and a
smaller $2.3\times$ gap on the high-rank Supernova), confirming that the
shortfall reflects model capacity rather than the density mismatch.

\begin{table}[t]
\centering\small
\caption{DBF at its matched-density upper bound ($\rho = 3\%$ for both
training and testing, no missing frames) versus our framework evaluated
at the same $\rho = 3\%$.}
\label{tab:dbf_matched}
\begin{tabular}{lccc}
\toprule
Full-availability RMSE & Ocean & Active Matter & Supernova \\
\midrule
DBF (matched $\rho=3\%$) & $0.346$ & $0.473$ & $0.681$ \\
TRACE-Smoother (ours)              & $0.040$ & $0.066$ & $0.298$ \\
ratio                    & $8.8\times$ & $7.1\times$ & $2.3\times$ \\
\bottomrule
\end{tabular}
\end{table}

\subsection{Streaming Filters That Failed to Reproduce}
We also attempted score-based ensemble filters, in both their native and
latent forms.

\textbf{EnSF} is training-free but, lacking a learned field prior,
collapses under the curse of dimensionality when run directly in the
field space ($d=65{,}536$ on Active Matter, $d=14{,}440$ on Ocean):
the ensemble cannot represent the posterior and the update suffers
frequent numerical breakdown (NaN). Enlarging the ensemble from $50$ to
$2{,}000$ particles does not help---RMSE stays at $\approx1.15$ on Active
Matter ($1.147\!\to\!1.225$) and $\approx1.03$ on Ocean, i.e.\ no better
than the trivial mean-field reconstruction (field std $\approx1.0$).
This is consistent with the EnSF design assumption of strong,
low-effective-rank PDE structure (e.g.\ Lorenz-96); we report it as an
empirical reproduction failure rather than a definitive diagnosis of its
cause.

\textbf{LD-EnSF} instead assimilates in a learned latent space whose
dynamics are a neural ODE (an LDNet); this learned forward model failed
to fit our data. On Active Matter the LDNet validation relative error
stagnated at $\approx\!1.0$ (no better than the mean field) after
training on $855$ trajectories, and in a controlled sanity check it could
not overfit even two trajectories (relative error plateaued at
$\approx\!0.71$). Without a usable latent forward model its assimilation
stage cannot be run. We therefore exclude both filters from the main
comparison and report them here for completeness.


\section{Additional Experimental Results}
\label{app:extra}\label{app:exp}

\begin{table*}[t]
\centering\small
\caption{Control regime (frame-rich: every frame observes the full domain
at density $\rho$), field-domain RMSE. Lower is better; \textbf{bold} is
best per column.}
\label{tab:control}
\begin{tabular}{l cc cc cc}
\toprule
 & \multicolumn{2}{c}{Active Matter} & \multicolumn{2}{c}{Ocean} & \multicolumn{2}{c}{Supernova} \\
\cmidrule(lr){2-3}\cmidrule(lr){4-5}\cmidrule(lr){6-7}
 & $1\%$ & $3\%$ & $1\%$ & $3\%$ & $1\%$ & $3\%$ \\
\midrule
LRTFR & 0.156 & 0.090 & 0.161 & 0.114 & 0.406 & 0.418 \\
MMGN  & 0.162 & 0.148 & 0.070 & 0.047 & 0.318 & 0.313 \\
SDIFT & 0.313 & 0.220 & 0.112 & 0.078 & 0.379 & 0.342 \\
DBF   & 0.747 & 0.473 & 0.527 & 0.346 & 0.771 & 0.681 \\
\midrule
TRACE-Frame  & 0.078 & 0.066 & 0.045 & 0.040 & 0.322 & 0.309 \\
TRACE-Filter & 0.078 & 0.066 & 0.047 & 0.042 & 0.321 & 0.308 \\
TRACE-Smoother           & \textbf{0.078} & \textbf{0.066} & \textbf{0.044} & \textbf{0.040} & \textbf{0.310} & \textbf{0.298} \\
\bottomrule
\end{tabular}
\end{table*}

We use the main-text method names throughout: \textbf{TRACE-Frame}
(moment-matched $S{=}20$ per-frame posterior, no temporal prior),
\textbf{TRACE-Filter} (causal SS-GP filter, no smoothing), and
\textbf{TRACE-Smoother} (full model with RTS smoothing); baselines are
\textbf{LRTFR}, \textbf{MMGN}, \textbf{SDIFT}, and \textbf{DBF}. A dash
(\,---\,) marks cells a method cannot produce (TRACE-Frame on
unobserved frames) or that were not run. All numbers are field-domain
RMSE averaged over test trajectories.

\subsection{Control Setting: Dense-Time Full-Domain Sparse Observation}
\label{app:exp:control}

We begin with a \emph{control} regime in which every frame is active and
independently observed by an off-grid random sample over the whole
domain ($\obsreg=\spdom$ for all $n$, density $\rho$).
This setting serves as a calibration: with each frame already
individually informative, it bounds how much of the reconstruction
quality is attributable to the generative field prior versus to
cross-time accumulation.
Table~\ref{tab:control} reports field-domain RMSE at
$\rho\in\{1\%,3\%\}$.

\begin{table*}[t]
\centering\small
\caption{Missing pattern with gap $n{=}1$ (every other frame active),
field-domain RMSE at $\rho\in\{1\%,3\%\}$. \textbf{Bold} is best per
column.}
\label{tab:miss1}
\begin{tabular}{l cc cc cc}
\toprule
 & \multicolumn{2}{c}{Active Matter} & \multicolumn{2}{c}{Ocean} & \multicolumn{2}{c}{Supernova} \\
\cmidrule(lr){2-3}\cmidrule(lr){4-5}\cmidrule(lr){6-7}
 & $1\%$ & $3\%$ & $1\%$ & $3\%$ & $1\%$ & $3\%$ \\
\midrule
LRTFR & 0.235 & 0.178 & 0.175 & 0.136 & 0.460 & 0.424 \\
MMGN  & 0.187 & 0.178 & 0.073 & 0.055 & 0.317 & 0.313 \\
SDIFT & 0.416 & 0.403 & 0.154 & 0.147 & 0.409 & 0.370 \\
DBF   & 0.907 & 0.755 & 0.713 & 0.587 & 0.829 & 0.771 \\
\midrule
TRACE-Frame  & --- & --- & --- & --- & --- & --- \\
TRACE-Filter & 0.207 & 0.198 & 0.082 & 0.079 & 0.335 & 0.322 \\
TRACE-Smoother           & \textbf{0.158} & \textbf{0.146} & \textbf{0.055} & \textbf{0.051} & \textbf{0.312} & \textbf{0.301} \\
\bottomrule
\end{tabular}
\end{table*}

TRACE-Smoother achieves the best RMSE on all three datasets at both
observation densities, outperforming every offline reconstructor.
Decomposing the contribution of each component, the per-frame prior
(TRACE-Frame) alone already surpasses or matches all offline
baselines, indicating that the latent diffusion prior drives most of
the gain when frames are individually informative. Adding the causal
SS-GP filter offers only marginal improvement, as propagating an
already-accurate per-frame estimate may accumulate minor error; the
RTS smoother consistently corrects this and yields a small
but reliable improvement. The overall temporal contribution in this
dense-time regime is thus modest---precisely as expected, since
cross-time accumulation is most valuable when individual frames are
under-informative, as demonstrated in the structured sensing regimes
of the main text.

\subsection{Temporally Sparse Observations: Additional Results}
\label{app:exp:grids}

We supplement the main-text Temporally Sparse experiments
(Table~\ref{tab:structured}) along two axes: a \emph{shorter} gap
(Table~\ref{tab:miss1}) and a \emph{higher} observation density
(Table~\ref{tab:temporal_rho03}). Both confirm that the honest boundary
in Sec.~5.4 is confined to the longer unobserved intervals of
Table~\ref{tab:structured}.

\paragraph{Shorter gap (Miss-1).}
Table~\ref{tab:miss1} reports the Missing pattern with gap $n{=}1$
(every other frame active) on all three datasets at
$\rho\in\{1\%,3\%\}$. TRACE-Smoother attains the lowest RMSE in every column.


\paragraph{Higher density ($\rho=3\%$).}
Table~\ref{tab:temporal_rho03} repeats the main-text
Table~\ref{tab:structured} grid at $\rho=3\%$. The ranking is preserved
across datasets, confirming that the main-text conclusions are stable
with respect to observation density.

\begin{table*}[t]
\centering\small
\caption{Temporally sparse observations at $\rho=3\%$, field-domain
RMSE. Columns follow main-text Table~\ref{tab:structured}
(\emph{Miss}: gap $n{=}3$; \emph{Blk-$L$}: blackout length $L$).
\textbf{Bold} is best per column.}
\label{tab:temporal_rho03}
\begin{tabular}{l ccc ccc ccc}
\toprule
 & \multicolumn{3}{c}{Active Matter} & \multicolumn{3}{c}{Ocean} & \multicolumn{3}{c}{Supernova} \\
\cmidrule(lr){2-4}\cmidrule(lr){5-7}\cmidrule(lr){8-10}
 & Miss & Blk-5 & Blk-10 & Miss & Blk-5 & Blk-10 & Miss & Blk-5 & Blk-10 \\
\midrule
LRTFR & 0.377 & 0.221 & 0.442 & 0.177 & 0.128 & 0.142 & 0.450 & 0.425 & 0.445 \\
MMGN  & 0.307 & 0.205 & 0.352 & \textbf{0.084} & 0.060 & \textbf{0.075} & \textbf{0.314} & 0.313 & \textbf{0.316} \\
SDIFT & 0.478 & 0.366 & 0.526 & 0.148 & 0.110 & 0.139 & 0.399 & 0.356 & 0.397 \\
DBF   & 0.901 & 0.617 & 0.742 & 0.747 & 0.502 & 0.627 & 0.829 & 0.747 & 0.812 \\
\midrule
TRACE-Frame  & --- & --- & --- & --- & --- & --- & --- & --- & --- \\
TRACE-Filter & 0.426 & 0.200 & 0.376 & 0.190 & 0.104 & 0.254 & 0.377 & 0.364 & 0.530 \\
TRACE-Smoother           & \textbf{0.285} & \textbf{0.173} & \textbf{0.321} & 0.093 & \textbf{0.056} & 0.121 & 0.323 & \textbf{0.300} & 0.349 \\
\bottomrule
\end{tabular}
\end{table*}

\subsection{Ablation: Prior Contribution and Fusion Variants}
\label{app:exp:ablation}

Table~\ref{tab:prior} isolates two contributions under full per-frame
observation. The $S{=}1\!\to\!20$ gap quantifies the
multi-sample moment-estimation gain of the Gaussian moment summary; the further
gap to TRACE-Smoother reflects the modest temporal contribution in the Control
regime (Sec.~5.3). Temporally-sparse behaviour of the same three
variants is in Table~\ref{tab:structured} (main text) and
Table~\ref{tab:miss1}.

\begin{table}[t]
\centering\small
\caption{Effect of the transition prior under full observation
(Control regime, $\rho=1\%$), field-domain RMSE. ``DPS ($S{=}1$)'' is
vanilla single-sample DPS.}
\label{tab:prior}
\begin{tabular}{lccc}
\toprule
 & Active Matter & Ocean & Supernova \\
\midrule
DPS ($S{=}1$)        & $0.091$ & $0.055$ & $0.365$ \\
TRACE-Frame        & $0.078$ & $0.045$ & $0.322$ \\
TRACE-Smoother                 & $\mathbf{0.078}$ & $\mathbf{0.044}$ & $\mathbf{0.310}$ \\
\bottomrule
\end{tabular}
\end{table}

\begin{table}[t]
\centering\small
\caption{Per-frame fusion time (Active Matter, Control regime,
$\rho{=}1\%$, $d{=}2304$; CPU, excluding DPS sampling).}
\label{tab:fusion}
\begin{tabular}{lcc}
\toprule
Fusion & Time/frame & Outcome \\
\midrule
Diagonal (ours)  & $1.7$\,ms        & RMSE $0.078$ \\
Full-covariance  & $\gtrsim\!20$\,s & OOM at $d{>}5000$ \\
\bottomrule
\end{tabular}
\end{table}

\begin{table}[t]
\centering\small
\caption{Sample-count $S$ ablation for TRACE-Frame on Active Matter
(Control regime, $\rho=1\%$, $10$ trajectories), field-domain RMSE.}
\label{tab:S_ablation}
\begin{tabular}{lcccc}
\toprule
$S$ & $1$ & $5$ & $20$ & $50$ \\
\midrule
RMSE & $0.091$ & $0.083$ & $\mathbf{0.081}$ & $0.080$ \\
\bottomrule
\end{tabular}
\end{table}

Table~\ref{tab:fusion} compares diagonal against full-covariance fusion
on Active Matter, timing the closed-form step alone. The diagonal
variant is $\mathcal{O}(d)$ in compute and memory, costs $1.7$\,ms per
frame, and runs on all $28$ trajectories. The full-covariance variant
forms and inverts a dense $d\times d$ precision matrix
($\mathcal{O}(d^3)$ compute, $\mathcal{O}(d^2)$ memory), is
$\gtrsim\!20$\,s per frame at $d{=}2304$, and diverges on $2$ of $5$
trajectories: the rank-$S{-}1{=}19$ empirical covariance from
$S{=}20$ samples is near-singular in the $d{=}2304$ space, so even
ridge-regularised inversion is numerically unstable. At the Supernova
scale ($d{=}32768$) the $\approx\!4$\,GB dense covariance exhausts
memory outright. This motivates the diagonal default; see
Appendix~\ref{app:diag-bias} for the rationale.

\paragraph{Posterior Gaussian summary: sample count $S$.}
Table~\ref{tab:S_ablation} sweeps $S$ on Active Matter (Control,
$\rho=1\%$): RMSE plateaus from $S{\ge}20$, justifying the default
used throughout (Appendix~\ref{app:impl}).

\begin{figure}[t]
\centering
\begin{minipage}[t]{0.49\linewidth}\centering
  \includegraphics[width=\linewidth]{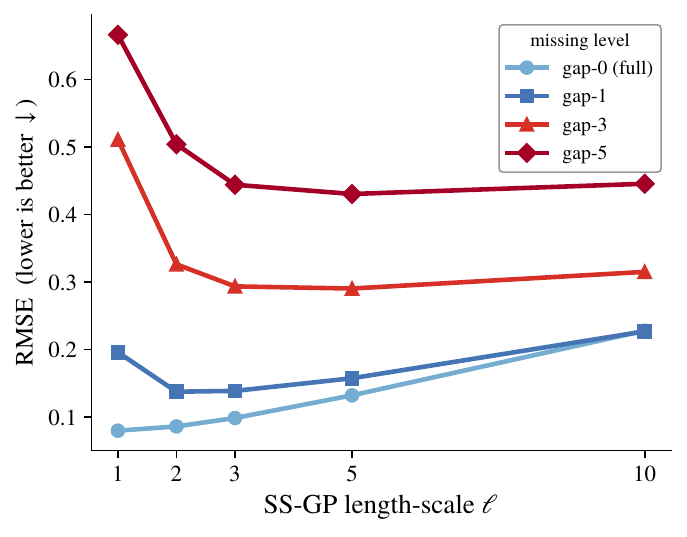}\\[1pt]
  {\small (a) Length-scale $\ell$ (fixed $\alpha\!=\!0.5$)}
\end{minipage}\hfill
\begin{minipage}[t]{0.49\linewidth}\centering
  \includegraphics[width=\linewidth]{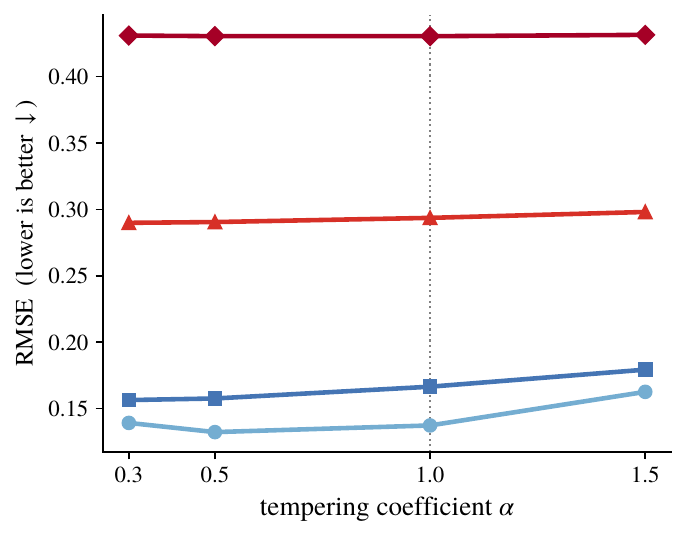}\\[1pt]
  {\small (b) Tempering $\alpha$ (fixed $\ell\!=\!5$)}
\end{minipage}
\caption{Hyperparameter sensitivity on Active Matter ($\rho=1\%$, full
TRACE-Smoother), one curve per Miss level $n\in\{0,1,3,5\}$.
\textbf{(a)} RMSE is strongly $\ell$-sensitive and the optimal $\ell$
shifts with $n$, saturating at $\ell\!\approx\!5$ for all sparse regimes.
\textbf{(b)} RMSE is nearly $\alpha$-insensitive (note the compressed
vertical range); the dotted line is the untempered Kalman update
($\alpha\!=\!1$). Other hyperparameters held fixed ($\sigma_f\!=\!1$,
diagonal fusion).}
\label{fig:sens}
\end{figure}

\begin{table}[t]
\centering\small
\caption{Observation-noise robustness (Active Matter, Control regime,
$\rho=1\%$), field-domain RMSE. $\sigma$ is the additive Gaussian sensor
noise. $\sigma{=}0$: $28$ trajectories; $\sigma{>}0$: $3$ trajectories.}
\label{tab:noise}
\begin{tabular}{lccc}
\toprule
$\sigma$ & TRACE-Frame & TRACE-Filter & TRACE-Smoother \\
\midrule
$0.0$ & $0.078$ & $0.078$ & $\mathbf{0.078}$ \\
$0.1$ & $0.097$ & $0.097$ & $0.098$ \\
$0.3$ & $0.195$ & $0.192$ & $\mathbf{0.184}$ \\
\bottomrule
\end{tabular}
\end{table}

\subsection{Sensitivity: Tempering and Lengthscale}
\label{app:exp:sensitivity}

TRACE-Smoother exposes two SS-GP hyperparameters: the Matérn length-scale $\ell$
and the tempering coefficient $\alpha$. Figure~\ref{fig:sens} sweeps
each on Active Matter ($\rho=1\%$) across four Miss levels $n\in\{0,1,3,5\}$
($n{=}0$ is Control, $n{=}1$ is Miss-1 of Table~\ref{tab:miss1},
$n{=}3$ is the main-text Miss) with TRACE-Smoother, holding
the other fixed.

\textbf{Length-scale (left).} Each curve is sharply U-shaped, with the
optimum shifting with the missing level: full observation ($n{=}0$)
prefers $\ell\!\approx\!1$, whereas all sparse regimes ($n{\ge}1$)
flatten into a common plateau at $\ell\!\approx\!5$. A single $\ell\!=\!5$ is
therefore near-optimal across sparse regimes, set from the
\emph{observable} missing rate rather than from oracle knowledge of
the target.

\textbf{Tempering (right).} Fixing $\ell\!=\!5$ and sweeping $\alpha$
leaves every curve nearly flat: RMSE varies by under $3\%$ over
$\alpha\in[0.3,1.0]$ and degrades only mildly at $\alpha\!=\!1.5$.
The dotted line at $\alpha\!=\!1$ marks the untempered Kalman update,
which differs marginally from the tempered default; tempering is a
gentle safeguard rather than the origin of the gains. We fix
$\alpha\!=\!0.5$ throughout.

Table~\ref{tab:obsmiss} decomposes the total RMSE of
Table~\ref{tab:structured} into averages over \emph{observed} frames
($\Obs_n\neq\emptyset$) and \emph{missing} frames
($\Obs_n=\emptyset$) under Miss-3, Blk-5, and Blk-10 at $\rho=1\%$.

\begin{table*}[t]
\centering\footnotesize
\caption{Per-frame error decomposition at $\rho=1\%$, stacked by dataset
(Active Matter, Ocean, Supernova). Each cell reports total /
observed-frame / missing-frame RMSE under Miss-3, Blk-5, and Blk-10
(same symbols as Table~\ref{tab:structured}). \textbf{Bold} marks the
best total per column within each dataset block.}
\label{tab:obsmiss}
\begin{tabular}{lccc}
\toprule
Method & Miss-3 & Blk-5 & Blk-10 \\
\midrule
\multicolumn{4}{c}{\textit{Active Matter} ($\rho{=}1\%$)} \\
\midrule
LRTFR           & $0.371 / 0.220 / 0.422$ & $0.218 / 0.156 / 0.451$ & $0.368 / 0.160 / 0.660$ \\
MMGN            & $0.313 / 0.156 / 0.366$ & $0.217 / 0.164 / 0.418$ & $0.360 / 0.164 / 0.635$ \\
SDIFT           & $0.458 / 0.363 / 0.490$ & $0.352 / 0.280 / 0.626$ & $0.472 / 0.303 / 0.710$ \\
DBF             & $0.930 / 0.826 / 0.965$ & $0.789 / 0.744 / 0.957$ & $0.843 / 0.749 / 0.975$ \\
TRACE-Filter  & $0.429 / 0.084 / 0.544$ & $0.211 / 0.103 / 0.619$ & $0.385 / 0.103 / 0.780$ \\
TRACE-Smoother            & $\mathbf{0.290} / 0.093 / \mathbf{0.356}$ & $\mathbf{0.186} / 0.132 / \mathbf{0.392}$ & $\mathbf{0.332} / 0.131 / \mathbf{0.612}$ \\
\midrule
\multicolumn{4}{c}{\textit{Ocean} ($\rho{=}1\%$)} \\
\midrule
LRTFR           & $0.201 / 0.194 / 0.203$ & $0.159 / 0.154 / 0.180$ & $0.172 / 0.162 / 0.187$ \\
MMGN            & $0.097 / 0.070 / 0.106$ & $0.080 / 0.070 / 0.119$ & $\mathbf{0.092} / 0.070 / \mathbf{0.123}$ \\
SDIFT           & $0.150 / 0.122 / 0.159$ & $0.113 / 0.098 / 0.168$ & $0.133 / 0.107 / 0.170$ \\
DBF             & $0.797 / 0.628 / 0.853$ & $0.604 / 0.536 / 0.862$ & $0.680 / 0.542 / 0.874$ \\
TRACE-Filter  & $0.191 / 0.045 / 0.239$ & $0.108 / 0.044 / 0.349$ & $0.257 / 0.045 / 0.554$ \\
TRACE-Smoother            & $\mathbf{0.095} / 0.045 / \mathbf{0.112}$ & $\mathbf{0.059} / 0.044 / \mathbf{0.116}$ & $0.124 / 0.045 / 0.235$ \\
\midrule
\multicolumn{4}{c}{\textit{Supernova} ($\rho{=}1\%$)} \\
\midrule
LRTFR           & $0.560 / 0.555 / 0.562$ & $0.430 / 0.437 / 0.414$ & $0.503 / 0.511 / 0.498$ \\
MMGN            & $\mathbf{0.318} / 0.315 / \mathbf{0.319}$ & $0.318 / 0.319 / 0.314$ & $\mathbf{0.320} / 0.320 / \mathbf{0.319}$ \\
SDIFT           & $0.435 / 0.420 / 0.440$ & $0.397 / 0.392 / 0.409$ & $0.424 / 0.407 / 0.433$ \\
DBF             & $0.853 / 0.796 / 0.871$ & $0.806 / 0.778 / 0.866$ & $0.841 / 0.782 / 0.877$ \\
TRACE-Filter  & $0.389 / 0.321 / 0.411$ & $0.375 / 0.318 / 0.500$ & $0.539 / 0.322 / 0.669$ \\
TRACE-Smoother            & $0.334 / 0.316 / 0.339$ & $\mathbf{0.312} / 0.310 / \mathbf{0.314}$ & $0.358 / 0.312 / 0.385$ \\
\bottomrule
\end{tabular}
\end{table*}

\begin{table*}[t]
\centering\small
\caption{Reliability on Active Matter (Control regime, $28$
trajectories): empirical coverage at two nominal credible levels, and
expected calibration error (ECE) over ten levels. Closer to the
nominal level is better.}
\label{tab:calib}
\begin{tabular}{lccc}
\toprule
 & nominal $0.90$ & nominal $0.95$ & ECE \\
\midrule
TRACE-Smoother             & $0.80$ & $0.86$ & $\mathbf{0.058}$ \\
TRACE-Frame ens.\ & $0.56$ & $0.62$ & $0.220$ \\
\bottomrule
\end{tabular}
\end{table*}

\subsection{Robustness: Observation Noise and Sparsity}
\label{app:exp:robust}

Table~\ref{tab:noise} reports robustness to additive measurement noise
on Active Matter ($\rho=1\%$, Control regime). All variants degrade
gracefully; TRACE-Smoother is most robust at the highest noise level
($0.184$ at $\sigma{=}0.3$), where the temporal prior helps most. At
low noise the three variants are indistinguishable.

\paragraph{Observation-rate sweep.}
Table~\ref{tab:rho_sweep} extends $\rho$ to $0.5\%$--$10\%$ on Active
Matter (Control). RMSE decreases monotonically with $\rho$ and
TRACE-Smoother remains on top of TRACE-Filter at every $\rho$, with
the largest margin at the sparsest setting.

\begin{table}[t]
\centering\small
\caption{Observation-rate sweep (Active Matter, Control regime,
$\sigma_{\text{obs}}=0$, $28$ trajectories, DPS reverse steps
$N{=}50$, guidance $\zeta\!\propto\!1/\rho$). Field-domain RMSE; lower
is better. This sweep uses a reduced $50$-step sampler and is therefore
not directly comparable to the default $100$-step results of
Table~\ref{tab:control}.}
\label{tab:rho_sweep}
\begin{tabular}{lcc}
\toprule
$\rho$ & TRACE-Filter & TRACE-Smoother \\
\midrule
$0.5\%$ & $0.104$ & $\mathbf{0.099}$ \\
$1\%$   & $0.093$ & $\mathbf{0.089}$ \\
$2\%$   & $0.082$ & $\mathbf{0.079}$ \\
$10\%$  & $0.072$ & $\mathbf{0.070}$ \\
\bottomrule
\end{tabular}
\end{table}

Table~\ref{tab:rho_sweep} extends $\rho$ to $0.5\%$--$10\%$ on Active
Matter (Control). RMSE decreases monotonically with $\rho$ and TRACE-Smoother
remains on top of TRACE-Filter at every $\rho$, with the largest
relative gain at the sparsest setting.

\subsection{Streaming Behaviour: Coverage Accumulation and Runtime}
\label{app:exp:stream}

\begin{table*}[t]
\centering\small
\caption{Coverage accumulation: RMSE versus number of revisits (loops)
under the circular moving window. ``TRACE-Smoother'' is the full model
(RTS smoothing); ``MMGN'' is the strongest offline per-frame baseline.}
\label{tab:coverage}
\begin{tabular}{lccccc}
\toprule
Loops & 1 & 2 & 3 & 4 & 6 \\
\midrule
AM, TRACE-Smoother & 0.537 & 0.498 & 0.476 & 0.475 & \textbf{0.468} \\
AM, MMGN           & 0.782 & 0.770 & 0.777 & 0.771 & 0.768 \\
\midrule
Ocean, TRACE-Smoother & 0.106 & 0.097 & \textbf{0.088} & 0.083 & 0.084 \\
Ocean, MMGN           & 0.277 & 0.283 & 0.271 & 0.284 & 0.286 \\
\midrule
SN, TRACE-Smoother & 0.334 & 0.320 & 0.318 & 0.314 & \textbf{0.311} \\
SN, MMGN           & 0.321 & 0.321 & 0.321 & 0.321 & 0.321 \\
\bottomrule
\end{tabular}
\end{table*}

\paragraph{Coverage accumulation.}
Table~\ref{tab:coverage} reports RMSE versus the number of sensor
revisits under the circular moving window (matching the Spatially
Localized setting of Table~\ref{tab:structured}). TRACE-Smoother improves
monotonically as coverage accumulates, while the per-frame INR
baseline is essentially flat in the revisit count. On Active Matter
and Ocean TRACE-Smoother dominates at every loop; on the high-rank Supernova
cube the baseline is competitive at a single pass but is overtaken
from the second revisit onward.

\begin{table*}[t]
\centering\small
\caption{Per-frame inference cost on Active Matter ($d=2304$), measured
on a single NVIDIA RTX~5090 GPU (the closed-form fusion step of
Table~\ref{tab:fusion} is timed on CPU).}
\label{tab:cost}
\begin{tabular}{lc}
\toprule
Component & Time \\
\midrule
DPS reverse ODE ($N{=}100$, $S{=}20$) & $\approx3.5$\,s\,/\,frame \\
SS-GP filter (forward) & $0.64$\,ms\,/\,frame \\
RTS smoothing (backward) & $2.35$\,ms\,/\,frame \\
FTM decode & $1.26$\,ms\,/\,frame \\
diagonal fusion ($d=32768$) & $4.1$\,ms\,/\,frame \\
\bottomrule
\end{tabular}
\end{table*}

\paragraph{Runtime and complexity.}
Table~\ref{tab:cost} reports per-frame cost on Active Matter
($d=2304$). The DPS reverse ODE ($\approx\!3.5$\,s) dominates; the
recursive filter / smoother / fusion / decode components add only a
few milliseconds. Diagonal fusion is $\mathcal{O}(d)$ in compute and
memory and remains feasible at $d{=}32768$ (Supernova), while
full-covariance fusion ($\mathcal{O}(d^3)$ compute,
$\mathcal{O}(d^2)$ memory) exhausts memory beyond $d\!\approx\!5000$.
The RTS backward pass adds one $\mathcal{O}(T)$ recursion at
$\approx\!3.5\times$ the forward filter cost; the DPS forward cost is
unchanged. Per-frame numbers extrapolate linearly in $T$ on our
validated $T=24$ horizon.


\subsection{Per-Frame Error Decomposition: Observed vs.\ Missing Frames}
\label{app:exp:obsmiss}

Two patterns emerge. On observed frames, TRACE-Filter and TRACE-Smoother
differ negligibly: direct per-frame DPS already constrains them. On
\emph{missing} frames, the smoother reduces error substantially over
the causal filter, especially under long blackouts---e.g.\ Active
Matter Blk-$10$ drops from $0.780$ to $0.612$ ($21\%$). This isolates
the role of the RTS backward pass: it improves precisely the frames
that the forward filter cannot constrain from later evidence.

\subsection{Uncertainty Calibration}
\label{app:exp:calib}

TRACE-Smoother produces a Gaussian posterior over the reconstructed field at
every frame. Table~\ref{tab:calib} reports empirical coverage against
nominal credible levels on Active Matter (Control regime, $28$
trajectories). Our posterior is substantially better calibrated than
the per-frame ensemble in magnitude (ECE $0.058$ vs.\ $0.220$), though it
retains some residual overconfidence at the higher credible levels. This is a magnitude-calibration check supporting the
probabilistic interpretation, not a claim about spatial
informativeness of the variance.


\subsection{Qualitative Reconstructions}
\label{app:exp:qual}

We provide per-dataset qualitative reconstructions for the structured
regimes (Temporally Sparse and Spatially Localized) at $\rho=1\%$.
Unless noted, columns are ground truth, observation, TRACE-Smoother,
TRACE-Filter, SDIFT, MMGN, DBF, LRTFR; the Observation panel shows
the scattered off-grid samples and is gray on unobserved frames. The
Control regime is not shown separately: its per-frame quality is
already visible in the observed-frame rows (marked \emph{obs}) of the
missing-frame figures.

\paragraph{Temporally sparse: missing frames.} The \emph{Miss} pattern
(one observed frame in four), matching the main-text Table~\ref{tab:structured}.
\begin{figure*}[t]
\centering
\includegraphics[width=0.85\textwidth]{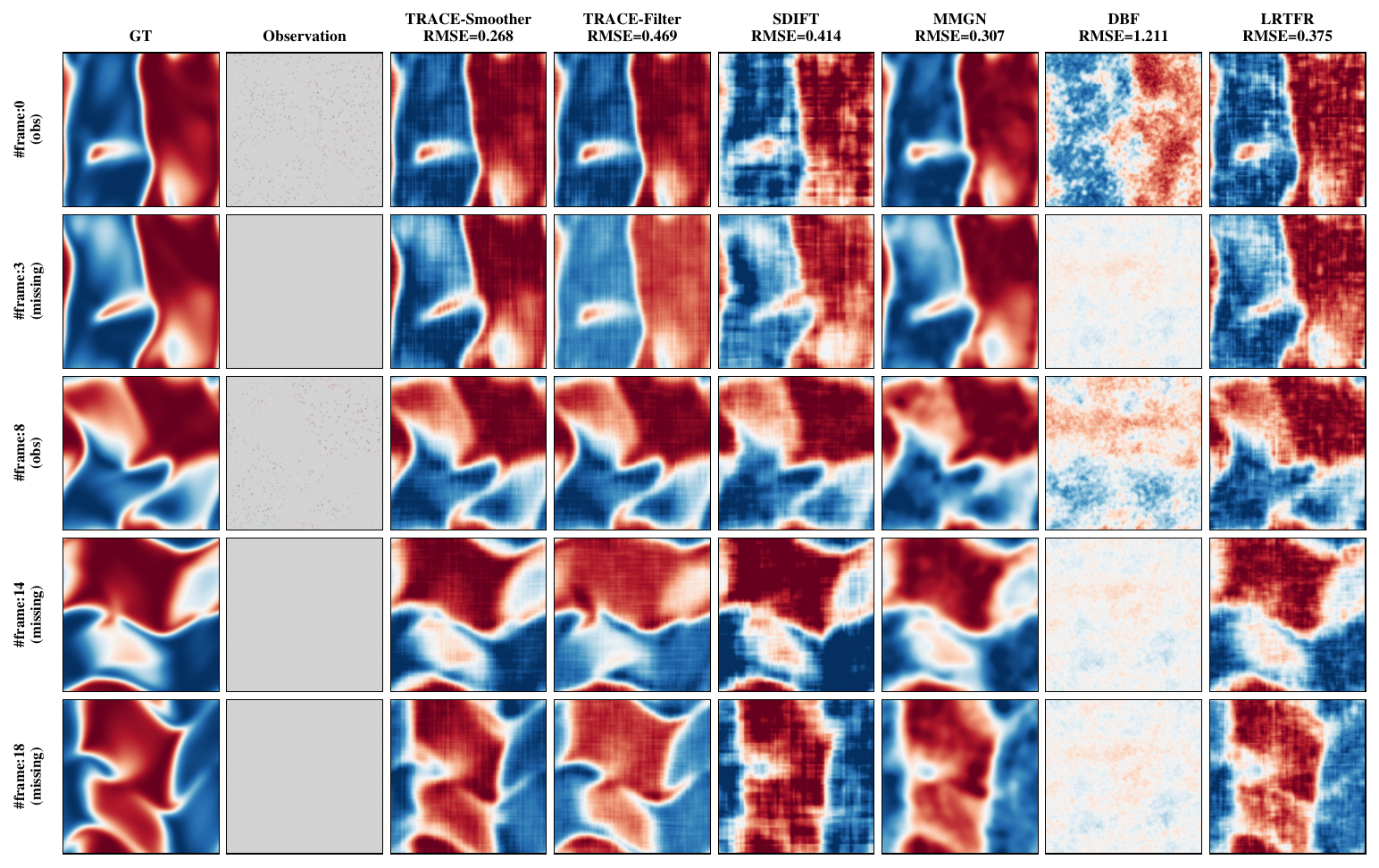}
\caption{Active Matter, \emph{Miss} pattern ($\rho=1\%$); five frames mixing observed and missing.}
\label{fig:qual_gap3_am}
\end{figure*}

\begin{figure*}[t]
\centering
\includegraphics[width=0.85\textwidth]{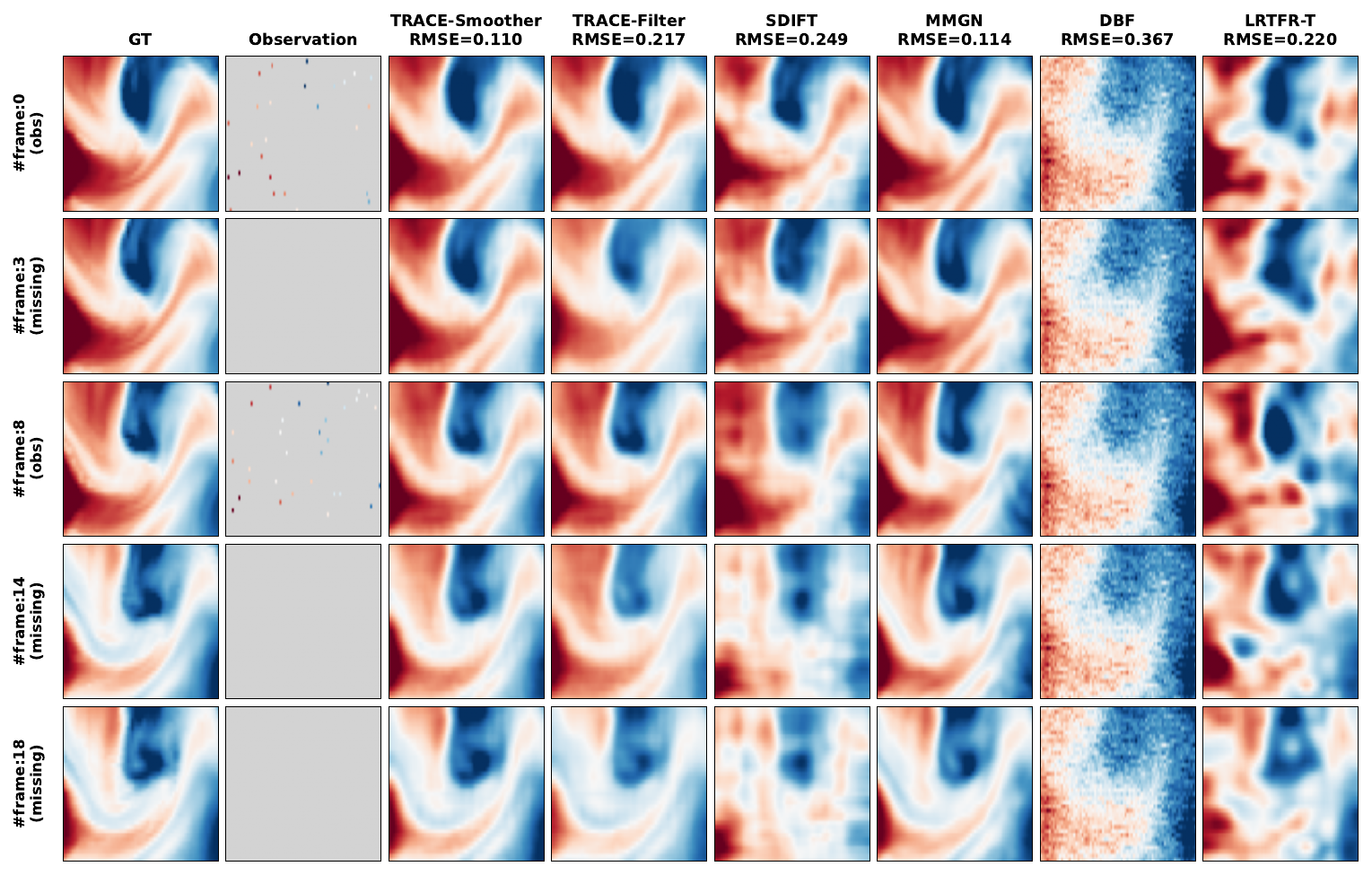}
\caption{Ocean, \emph{Miss} pattern ($\rho=1\%$, gap $n{=}3$): TRACE-Smoother recovers coherent spatial structures throughout missing intervals; TRACE-Filter drifts on gap frames; offline baselines produce temporally inconsistent transitions. Columns: ground truth, observation, TRACE-Smoother, TRACE-Filter, SDIFT, MMGN, DBF, LRTFR.}
\label{fig:qual_gap3_ocean}
\end{figure*}

\begin{figure*}[t]
\centering
\includegraphics[width=0.85\textwidth]{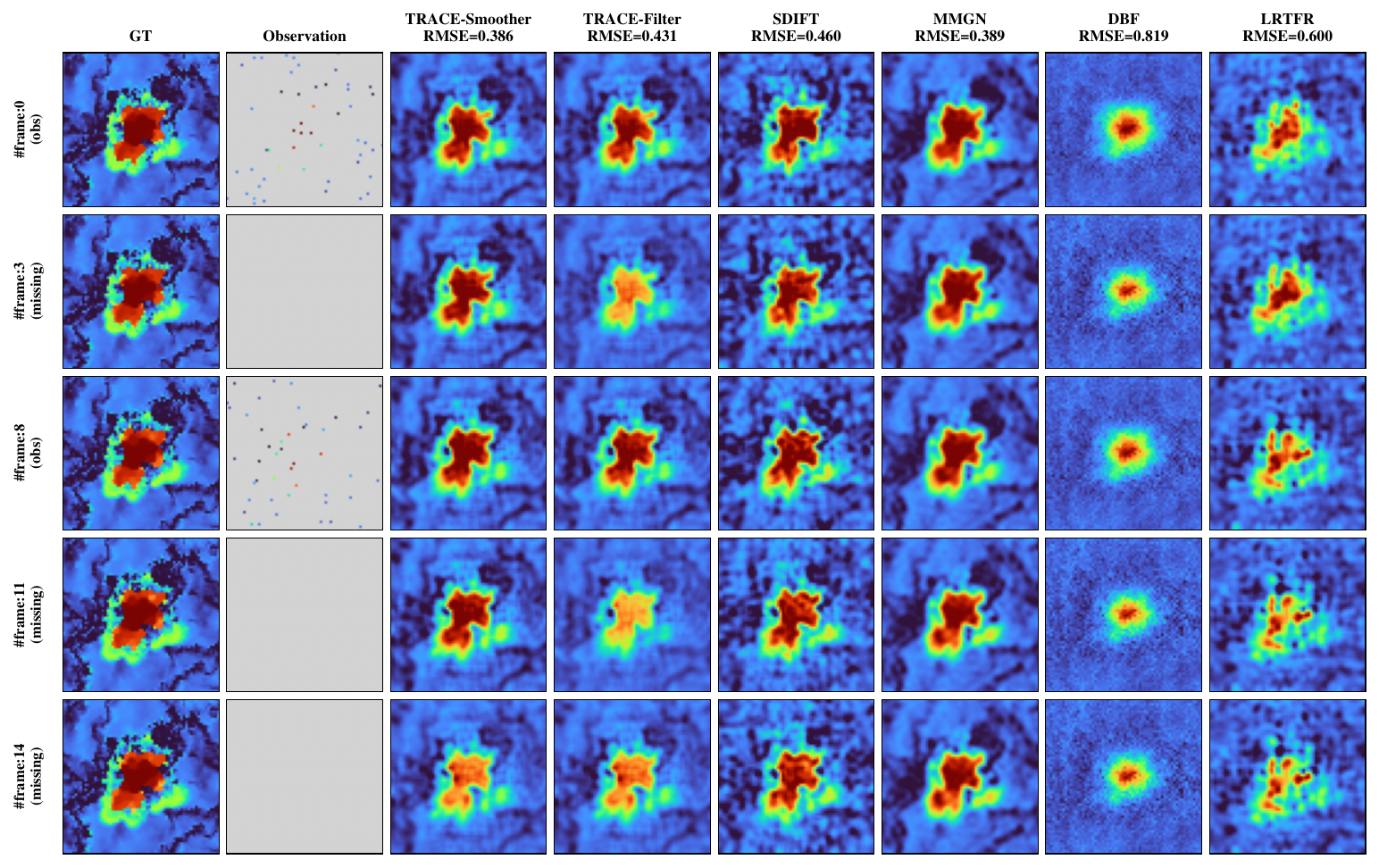}
\caption{Supernova ($z{=}32$ slice), \emph{Miss} pattern ($\rho=1\%$).}
\label{fig:qual_gap3_sn}
\end{figure*}

\paragraph{Temporally sparse: blackout.} A contiguous interval is
unobserved (gray in the Observation column); nine frames span the
blackout. The causal TRACE-Filter loses structure inside the
blackout, while TRACE-Smoother propagates evidence from both
ends to bridge it. The blackout (Blk-$10$) covers frames $7$--$16$
(Active Matter, Ocean) and $3$--$12$ (Supernova).
\begin{figure*}[t]
\centering
\includegraphics[width=\textwidth]{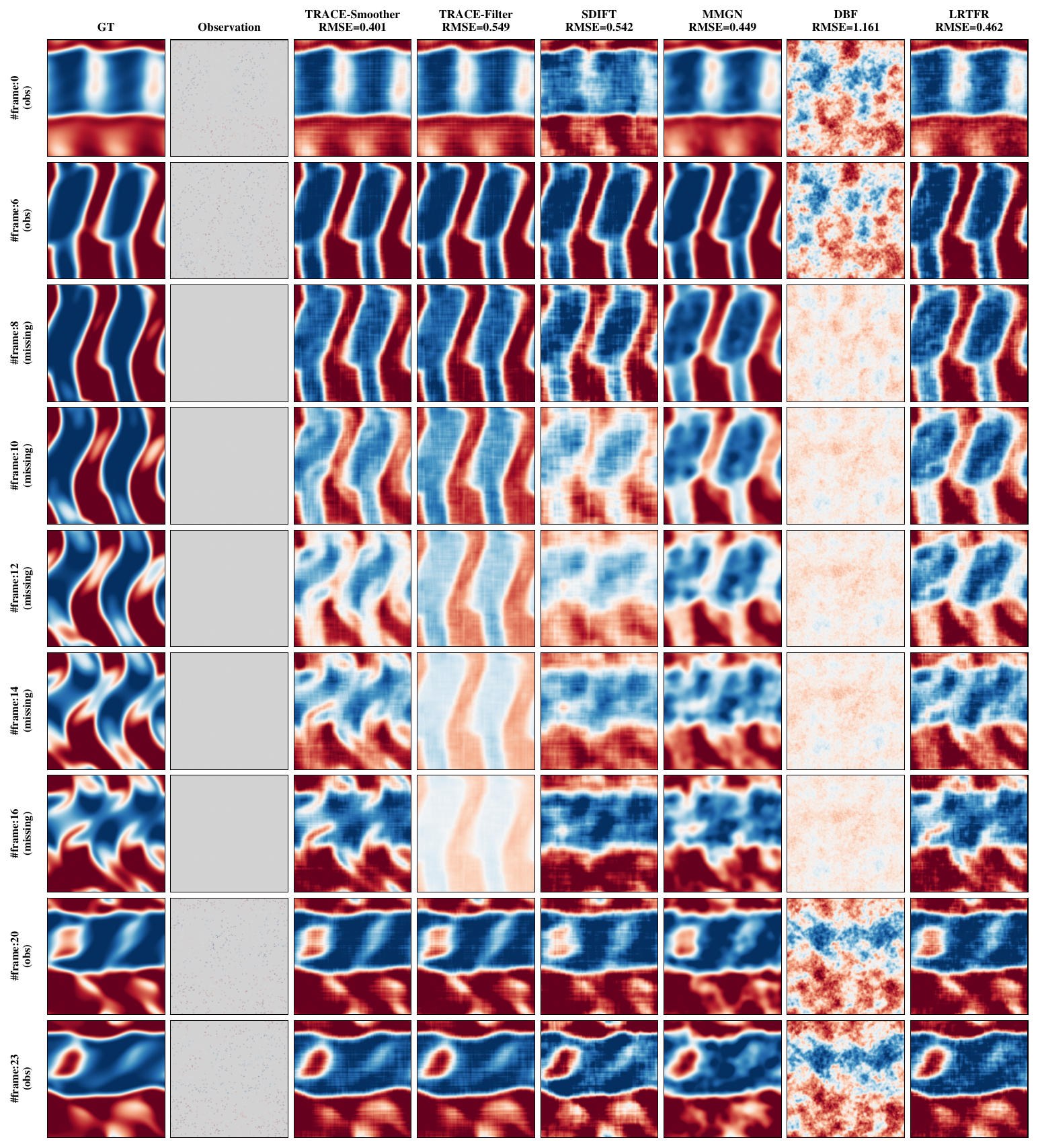}
\caption{Active Matter, Blk-$10$ ($\rho=1\%$): frames $7$--$16$ unobserved.}
\label{fig:qual_block10_am}
\end{figure*}

\begin{figure*}[t]
\centering
\includegraphics[width=\textwidth]{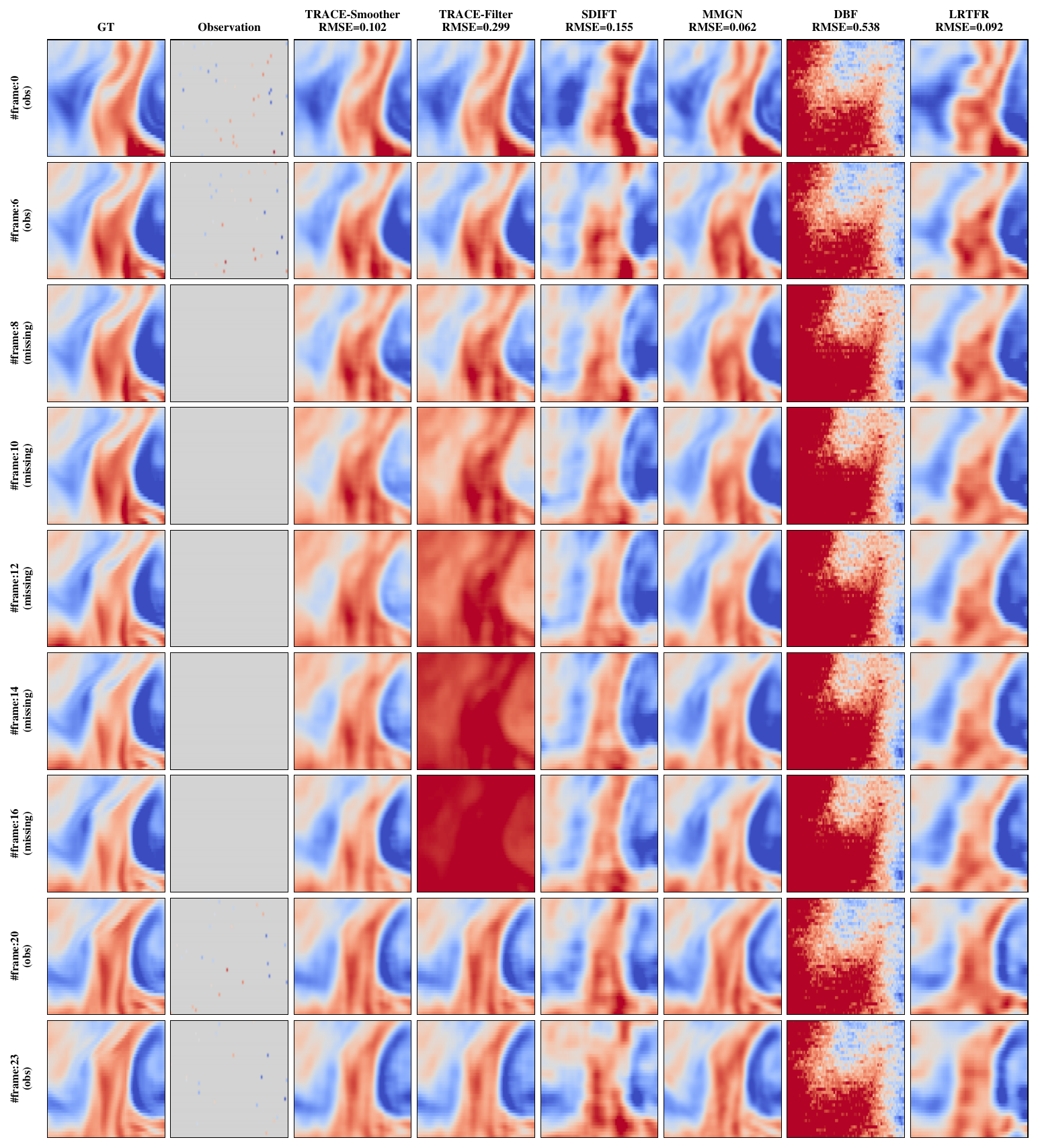}
\caption{Ocean (layer 2), Blk-$10$ ($\rho=1\%$): frames $7$--$16$ unobserved.}
\label{fig:qual_block10_ocean}
\end{figure*}

\begin{figure*}[t]
\centering
\includegraphics[width=\textwidth]{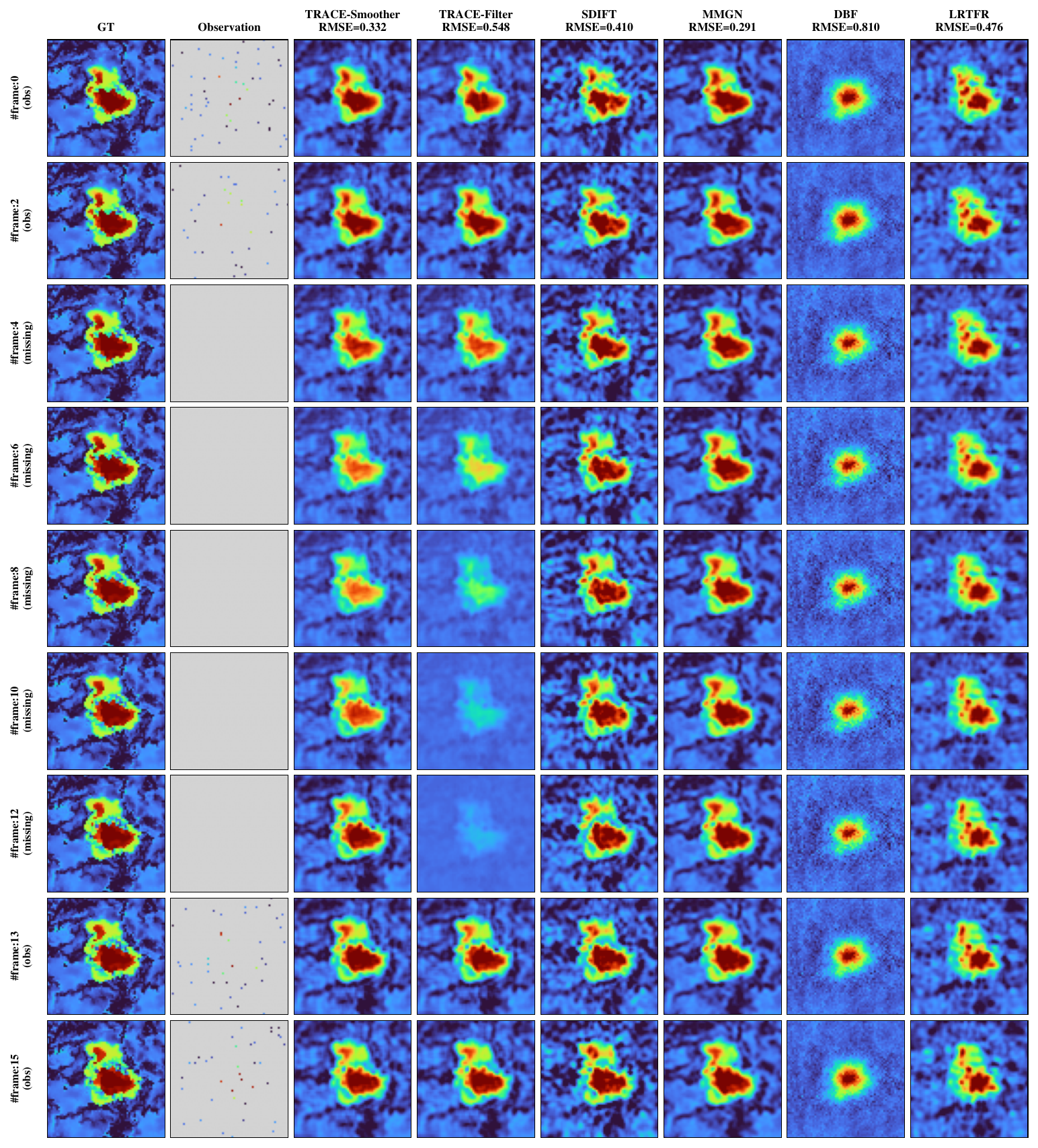}
\caption{Supernova ($z{=}32$), Blk-$10$ ($\rho=1\%$): frames $3$--$12$ unobserved.}
\label{fig:qual_block10_sn}
\end{figure*}

\paragraph{Spatially localized: moving window.} A circular one-lap
window (green box) sweeps the domain. TRACE-Smoother carries previously-swept
regions forward into a full-field estimate. Columns here are ground
truth, observation, TRACE-Smoother, TRACE-Filter, TRACE-Frame, SDIFT, MMGN,
LRTFR: the TRACE-Frame column exposes how the memoryless variant
collapses outside the current window; DBF is dropped (degenerates to a
near-mean field).
\begin{figure*}[t]
\centering
\includegraphics[width=\textwidth]{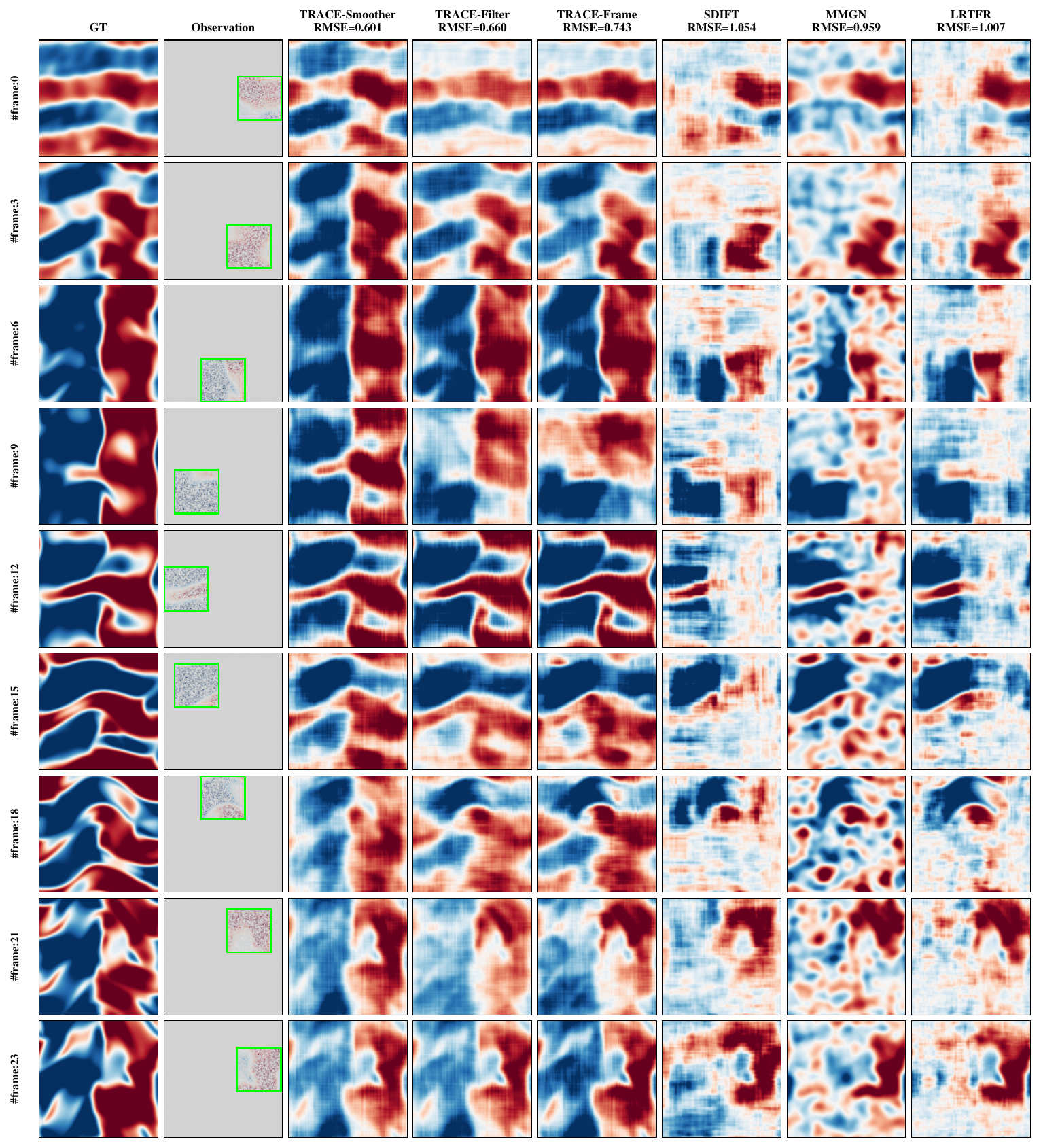}
\caption{Active Matter, moving window (circular, one lap, $\rho_{\text{loc}}=15\%$); nine frames span one sweep. TRACE-Smoother accumulates previously-swept regions into a coherent global field, while the per-frame baselines collapse outside the current window.}
\label{fig:qual_window_am}
\end{figure*}

\begin{figure*}[t]
\centering
\includegraphics[width=\textwidth]{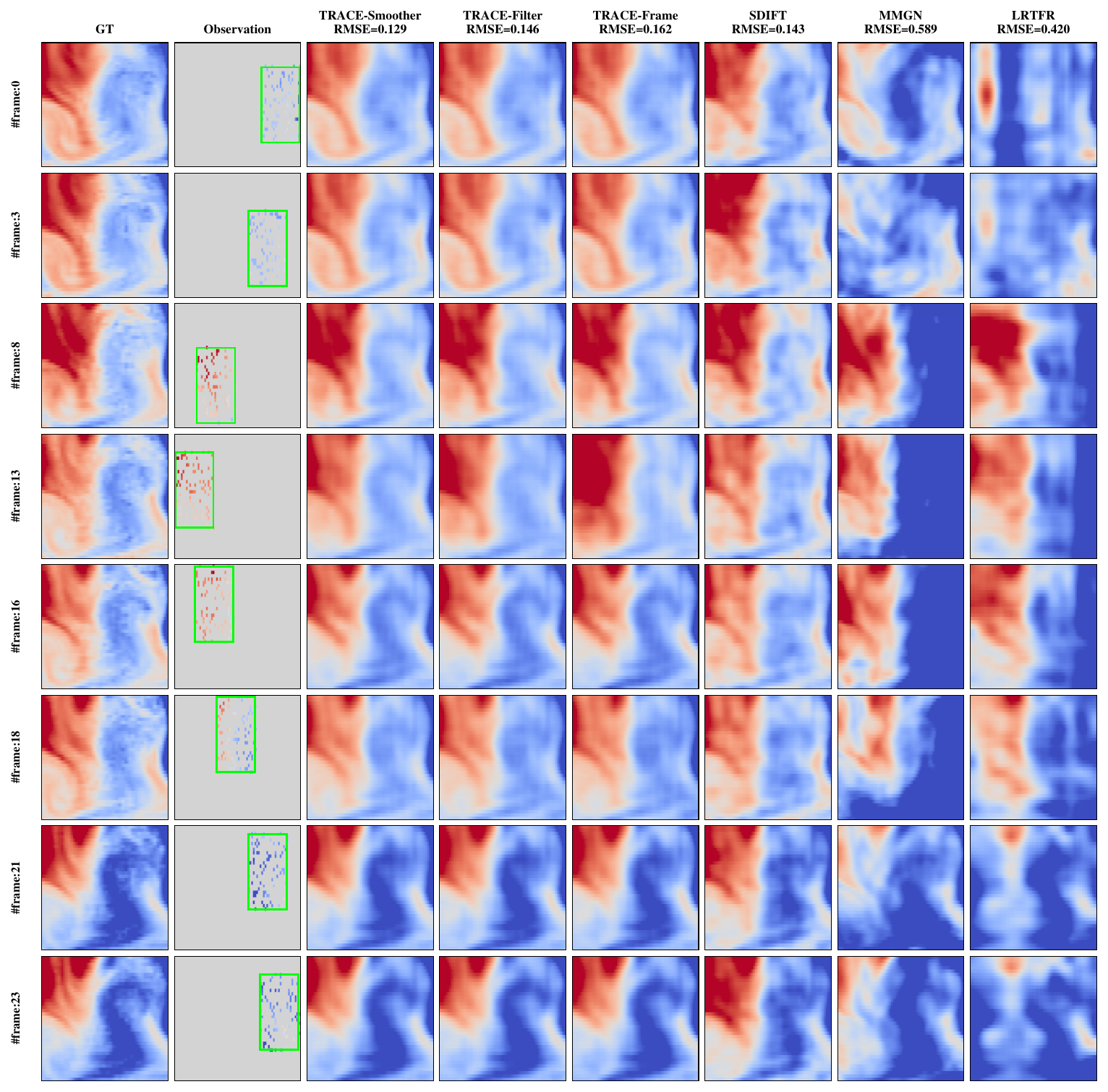}
\caption{Ocean (layer 2), moving window (circular, one lap, $\rho_{\text{loc}}=15\%$).}
\label{fig:qual_window_ocean}
\end{figure*}

\begin{figure*}[t]
\centering
\includegraphics[width=\textwidth]{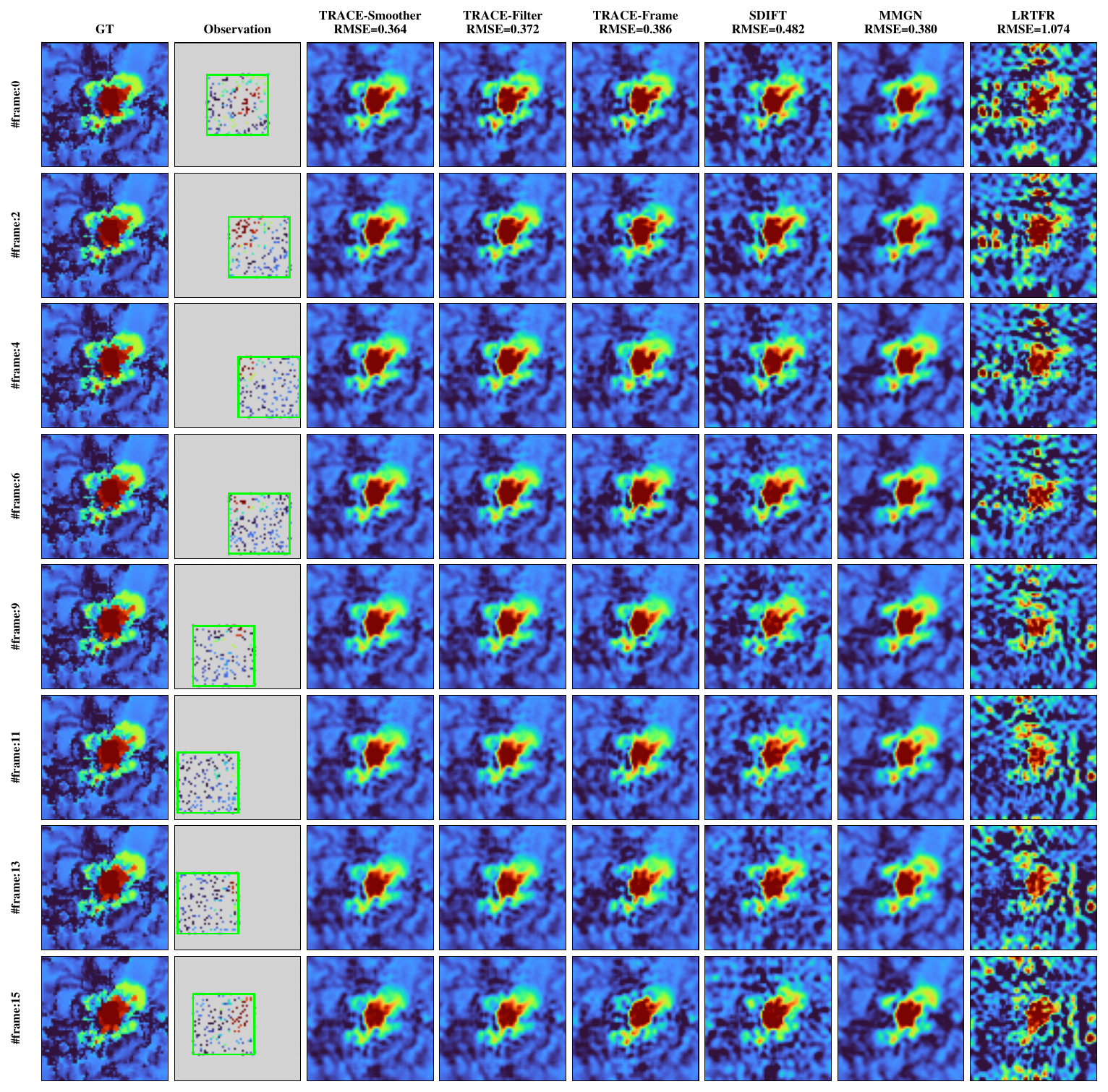}
\caption{Supernova ($z{=}32$), moving cube (circular, one lap).}
\label{fig:qual_window_sn}
\end{figure*}

\paragraph{Failure case.} When the narrow window leaves the field
under-determined, the learned generative prior cannot resolve it: the
causal TRACE-Filter propagates the erroneous per-frame estimate
forward and even the full smoother cannot recover---a limitation when
per-frame evidence is too weak for the prior to be informative.
\begin{figure*}[t]
\centering
\includegraphics[width=\textwidth]{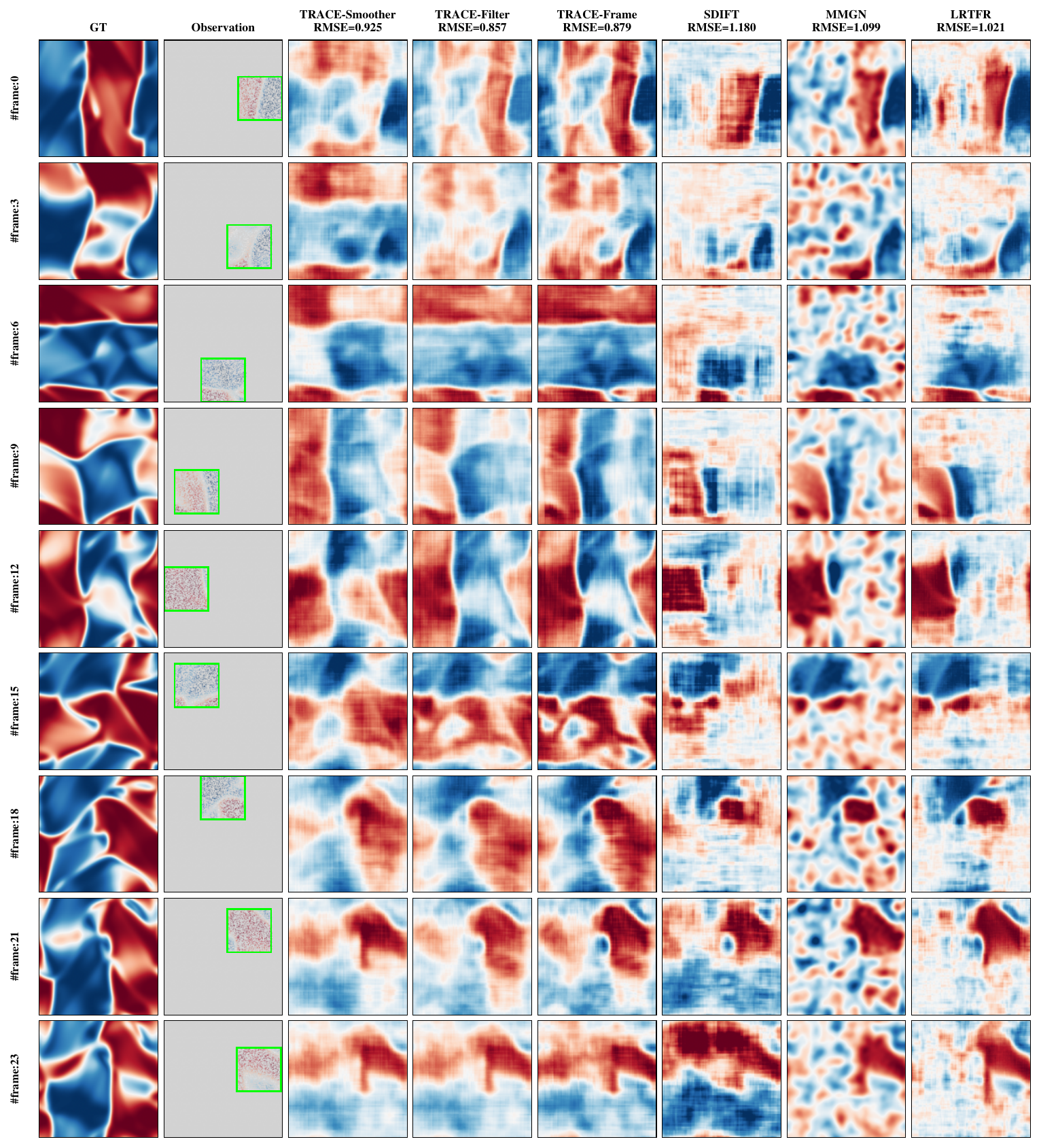}
\caption{Failure case: Active Matter moving window where per-frame evidence is too weak for the generative prior; TRACE-Filter propagates the error and TRACE-Smoother cannot recover.}
\label{fig:qual_fail}
\end{figure*}

\end{document}